\documentclass[11pt]{article}

\usepackage[final]{acl}

\usepackage{times}
\usepackage{latexsym}

\usepackage[T1]{fontenc}
\usepackage[utf8]{inputenc}

\usepackage{microtype}
\usepackage{inconsolata}

\usepackage{graphicx}

\usepackage{amsmath}
\usepackage{amssymb}
\usepackage{amsfonts}
\usepackage{nicefrac}

\usepackage{booktabs}
\usepackage{tabularx}
\usepackage{makecell}
\usepackage{multirow}
\usepackage{pifont} 
\usepackage{dblfloatfix}
\usepackage{cuted}

\usepackage{enumitem}
\usepackage{subcaption}
\usepackage[most]{tcolorbox}

\usepackage{CJKutf8}

\definecolor{promptbg}{gray}{0.95}
\definecolor{promptframe}{gray}{0.2}

\title{Adapting Technical-Service LLM Agents with Latent Logic Augmentation, Robust Noise Reduction, and Hybrid Reward Modeling}

\author{
\textbf{Junzhuo Ma$^{1,3,*}$, Chenghuang Shen$^{2,3,*}$, Yi Yu$^{1,3,*}$, Xingyan Liu$^{3,*}$, Jing Gu$^{1,3}$, Hangyi Sun$^{1,3}$,} \\
\textbf{Guangquan Hu$^{3,4}$, Jianfeng Liu$^{3,\dagger}$, Weiting Liu$^{3,4}$, Pu Mingyue$^{3}$, Wang Yu$^{3}$,} \\
\textbf{Zhengdong Xiao$^{3}$, Rui Xie$^{3}$, Longjiu Luo$^{3}$, Qianrong Wang$^{3}$, Gurong Cui$^{3}$,} \\
\textbf{Honglin Qiao$^{3}$, Wenlian Lu$^{1,2,4,5,6,\dagger}$} \\
$^{1}$School of Mathematical Sciences, Fudan University, Shanghai, China \\
$^{2}$Shanghai Center for Mathematical Sciences, Fudan University, Shanghai, China \\
$^{3}$Alibaba Group, Hangzhou, China \\
$^{4}$Institute of Science and Technology for Brain-Inspired Intelligence, Fudan University, Shanghai, China \\
$^{5}$Center for Applied Mathematics \& Shanghai Key Laboratory of Contemporary Applied Mathematics, \\
Fudan University, Shanghai, China \\
$^{6}$Ji Hua Laboratory, Foshan, 528200, China \\[3pt]
\parbox{0.98\textwidth}{\centering\small\normalfont\ttfamily
\{23110180025,y\_yu21,24210180034,24210180057,22110850002,wtliu24\}@m.fudan.edu.cn \\
\{chenghuang.sch,wxw,jiawei.ljf,pumingyue.pmy,xunzhen.wy,tianqiao.xzd,harry.xr, \\
xuancan,qianrong.wqr,yuanchuo,kenny.qhl\}@alibaba-inc.com;\quad wenlian@fudan.edu.cn}
}

\begin{document}
\maketitle
\begingroup
\renewcommand{\thefootnote}{}
\footnotetext{$^{*}$Equal contribution. $^{\dagger}$Corresponding authors.}
\endgroup

\begin{abstract}
  Technical-service LLM agents are entering production workflows, where value depends on whether engineers adopt generated replies. Service tickets hide decision logic, contain noisy single-reference responses, and make reward evaluation costly, making standard post-training brittle.
  Existing post-training and LLM-as-a-Judge approaches improve grounding or feedback, but do not jointly model latent decision logic, response diversity, and reward cost. We address this gap by coupling latent logic augmentation, robust noise reduction, and hybrid reward modeling. The framework augments supervised fine-tuning data with Planning-Aware Trajectory Modeling and Reasoning Augmentation, builds dual-filtered Multiple Ground Truths, and trains the policy with a hybrid reward that combines a Reranker with an LLM-as-a-Judge. On real Cloud technical-service tasks, the adapted Qwen3-4B achieves the highest Multi-ECS (0.441), lower reward cost, and the highest production adoption rate (46.63\%).
\end{abstract}

\section{Introduction}
\label{sec:intro}
\begin{figure*}[t]
  \centering
  \includegraphics[width=\textwidth]{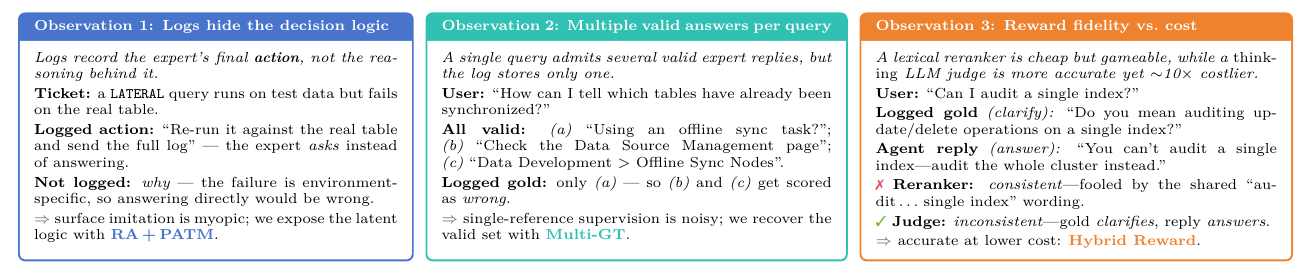}
  \caption{\textbf{Three observations that motivate our framework}, each grounded in a real Cloud technical-service ticket, mapping a pain point to the component that addresses it: hidden decision logic (\textbf{RA\,+\,PATM}), multiple valid answers for one query (\textbf{Multi-GT}), and a gameable cheap Reranker vs.\ an accurate costly judge (\textbf{Hybrid Reward}).}
  \label{fig:motivation}
\end{figure*}

Large Language Model (LLM) agents are increasingly used across web, tool-use, and Cloud-service workflows \cite{deng2023mind2web, zheng2024gpt, team2025kimi, zeng2025glm, team2025longcat, yu2026cirrusbench}. In our Cloud technical-service setting, practical value is measured by whether generated replies are adopted in production.
Inference-time adaptation through In-Context Learning (ICL) \cite{brown2020language, wei2022chain, reid2024gemini, agarwal2024many, zhang2025agentic} and Retrieval-Augmented Generation (RAG) \cite{lewis2020retrieval, cuconasu2024power, mialonaugmented} helps reuse domain context, but improving adoption rate in complex Cloud technical-service scenarios typically requires parameter updates. The standard post-training involves Supervised Fine-Tuning (SFT) and Reinforcement Learning (RL), which exposes several coupled obstacles in practice (Figure~\ref{fig:motivation}).

First, conventional training and interaction data often expose final outputs or actions without explicitly modeling the intermediate reasoning that produced them \cite{zelikman2022star, zelikmanquiet, li2026s}.
SFT, human preference alignment \cite{ouyang2022training}, Continual Pre-training \cite{su2025scaling}, and Parameter-Efficient Fine-Tuning (PEFT) \cite{hulora} can therefore encourage myopic imitation instead of transferable decision reasoning.
Second, single-reference supervision penalizes valid but different responses in semantically diverse service tasks, leading to output homogenization or ``knowledge collapse'' \cite{wright2025epistemic, shypula2025evaluating, jiang2025artificial}.
This is not resolved by stronger optimizers alone: although alignment methods, including RLHF, PPO, DPO, DAPO, and recent multi-reference, agent-RL, or multi-answer RL variants, can improve alignment \cite{ziegler2019fine, Schulman-PPO, rafailov2023direct, wu2025intelligently, aminian2025kl, yu2025dapo, xi2026rlgeneralization, yuan2026vpr, zhang2026qevolve, puri2026reaching}, their reward signals stay fragile when the space of valid responses is only partially represented.
Third, LLM-as-a-Judge reward modeling provides scalable semantic evaluation \cite{Liu-G-eval, Zheng-judging, Li-judges, yuan2025selfrewardinglanguagemodels, lee2023rlaif}, but it is vulnerable to reward hacking and costly for large-scale online adaptation \cite{anwar2024foundational, coste-reward, eisenstein-helping, xu2024perfect, saha2026judgebudget, zhang2026reasoningnotfree}.

We propose an adaptation framework that addresses these obstacles with three coordinated components.
\textbf{Latent Logic Augmentation} makes hidden decision structure explicit through \textit{Planning-Aware Trajectory Modeling (PATM)} and \textit{Reasoning Augmentation (RA)}, stabilizing SFT for planning-intensive tasks.
\textbf{Robust Noise Reduction} builds a \textit{Multiple Ground Truths (Multi-GT)} dataset through dual filtering, preserving distinct but valid solution paths.
\textbf{Hybrid Reward Modeling} (HRM)
combines
an LLM-as-a-Judge signal with a lightweight Reranker, retaining reward fidelity while reducing training cost.
We validate the framework on real-world Cloud technical-service tasks using ECS-based offline evaluation and production adoption rate, showing gains in stability, response diversity, and practical deployability. Matched Multi-GT DPO and adaptive judge-routing comparisons position the method against relevant alternatives. Transfer experiments on Llama-3.1-8B and Qwen3-32B test its generality across model sizes and families.

\section{Related Work}
\label{sec:relatedwork}
\textbf{LLM Adaptation and Latent Logic.}
LLM specialization spans inference-time methods such as ICL and RAG \cite{brown2020language, wei2022chain, lewis2020retrieval} and parameter-update methods such as SFT, RLHF, PPO, DPO, and DAPO \cite{ouyang2022training, ziegler2019fine, Schulman-PPO, rafailov2023direct, yu2025dapo}.
Recent agentic RL studies further examine generalization, dense process rewards, and self-evolving policy learning for interactive agents \cite{xi2026rlgeneralization, yuan2026vpr, zhang2026qevolve, li-etal-2026-rethinking}.
For technical-service agents, the central difficulty is not only optimization stability but also the missing logic in logged expert trajectories.
Prior work injects future dialogue context, latent thought traces, reusable skills, or interaction histories to improve planning and reasoning \cite{zeng2023futuretod, zelikman2022star, zelikmanquiet, song2024agentbank, huang2026skill}, often with environment-mediated feedback \cite{gandhi2024stream, lin2023learning, serl, liu2026toolanchor}.
Our Latent Logic Augmentation instead structures offline training data with PATM and RA so the model learns decision reasoning before RL.

\textbf{Diverse Valid Responses.}
Single-reference training and evaluation can mislabel valid alternatives as errors, especially when several replies can solve the same technical issue \cite{wright2025epistemic, shypula2025evaluating, liu-etal-2026-profit}.
Recent benchmarks and RL methods address this by introducing multiple references or multi-answer objectives \cite{jiang2025artificial, wu2025intelligently, aminian2025kl, yu2025dapo, puri2026reaching}.
Noise-aware structured reasoning similarly filters irrelevant question/table evidence or supervises pruning trajectories to preserve answer-critical content \cite{ye-etal-2026-tableqa, guo-etal-2026-rethinking-table}.
Our Robust Noise Reduction follows this direction but targets real-world service logs: dual filtering expands each logged response into a Multi-GT set used for both ECS evaluation and RL training.

\textbf{Judge-Based Reward Modeling.}
LLM-as-a-Judge methods provide scalable semantic evaluation and reward generation \cite{Liu-G-eval, Zheng-judging, Li-judges,
yuan2025selfrewardinglanguagemodels, lee2023rlaif,
wen-etal-2026-critic}.
However, judge rewards can be exploited by the policy, and ensemble or large-judge solutions increase online training cost \cite{ouyang2022training, anwar2024foundational, coste-reward,
xu2024perfect, huang2026realtimealignedrewardmodel,
zeng-etal-2026-tinyjudge}.
Recent budget-aware judge allocation, adaptive judge routing, and stepwise model routing work similarly highlights the need to spend expensive reasoning computation selectively \cite{saha2026judgebudget, zhang2026reasoningnotfree, ye2026rubricguidedprocessrewardstepwise, ye2026rethinkingstepwisemodelrouting}.
Our HRM addresses this issue by combining a strong LLM-as-a-Judge signal with a lightweight Reranker, reserving expensive judgment for cases where the fast signal is less reliable.

\section{Methodology}
\label{sec:methodology}

\begin{figure*}[htbp]
    \centering
    \includegraphics[width=\textwidth]{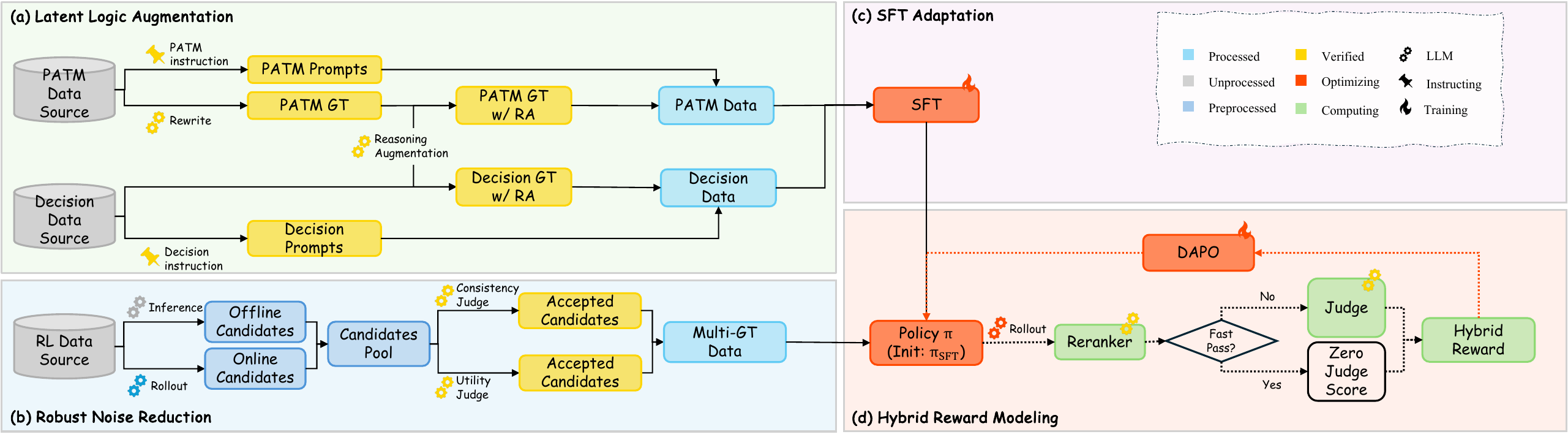}
    \caption{\textbf{Overview of the Proposed Framework.} The framework consists of four stages: (a) Latent Logic Augmentation; (b) Multi-GT data construction; (c) SFT adaptation; (d) RL adaptation with HRM.
    }
    \label{fig:overall_framework}
\end{figure*}

In this section, we present our framework for adapting Large Language Models (LLMs) to complex technical-service domains~(Figure~\ref{fig:overall_framework}). We first establish a Multi-GT paradigm for evaluation and data construction. On this foundation, adaptation comprises two stages: an \textbf{SFT} stage whose data we enrich with \textbf{Latent Logic Augmentation} (PATM and RA) to surface latent decision logic, followed by an \textbf{RL} stage driven by HRM that fuses an LLM-as-a-Judge with a lightweight Reranker.


\subsection{Foundation: Multi-GT Data and Evaluation}
\label{sec:multi_gt_foundation}

The foundation of our framework is a Multi-GT paradigm designed to address the inherent ambiguity of technical-service tasks, where a single query can have multiple valid responses. Traditional evaluation, which uses a single logged agent response as the unique gold reference, is unreliable because it unfairly penalizes valid but distinct solutions. To overcome this, for a query $q$ we expand the single gold reference into a set of valid responses $\mathcal{Y}^\star(q)=\{y_1^\star,\dots,y_m^\star\}$ used for both evaluation and training.

\textbf{Core Instruments: Consistency Judge and Utility Judge.}
Since manual annotation is impractical, we automate the construction of the Multi-GT training set via a pre-computation pipeline. Central to our pipeline are two LLM-based judges:

(1) \textit{Consistency Judge} evaluates whether a response follows business logic consistent with the expert (see prompt in Appendix~\ref{consistency prompt}). Validated against 152 human expert annotations, it attains $91.5\%$ accuracy and ``almost perfect'' agreement with human labels ($\kappa=0.82$); full statistics (confidence intervals and benchmark comparisons) are reported in Appendix~\ref{app:judge_reliability}.

(2) \textit{Utility Judge} uses privileged context (summary of service ticket) to determine if an alternative response effectively resolves the customer's issue. It captures valid solutions that differ from the history and achieves 83\% alignment with human labels (see prompt in Appendix~\ref{utility judge}).

\textbf{Automated Construction via Dual-Filtering Expansion.}
\label{sec:diversity_expansion}
We construct candidates for Multi-GT dataset through two complementary streams:
(1) \textbf{Offline Exploration:} We use a lightweight model (e.g., Qwen3-4B) with high temperature ($T=1.2$) to generate diverse candidates. This injects novel linguistic patterns and reasoning angles; (2) \textbf{Online Adaptation:} We harvest high-likelihood rollouts from a preliminary RL run of a policy model. This stream exposes ``hard positive'' responses favored by the model that are functionally valid but differ from the human reference.

All candidates undergo a \textit{Dual-Filtering} process comprising a Consistency Judge and a Utility Judge: The Consistency Judge ensures policy adherence, while the Utility Judge validates the effectiveness of alternative solutions. Only candidates passing one of these filters are added to $\mathcal{Y}^\star(q)$.

For transparency, we report the detailed dataset composition and the expansion breakdown by source in Appendix~\ref{app:multigt_stats} (Table~\ref{tab:multigt_stats_grouped}). Overall, Multi-GT expansion roughly doubles the number of references (e.g., Train: 5{,}120 $\rightarrow$ 10{,}127). For the test split, every reference added by either the Consistency or Utility branch also passes human review, with per-branch audit statistics and raw-vs.-cleaned results provided in Appendix~\ref{app:multigt_stats}.

\textbf{Evaluation metric: Single- and Multi-ECS.}
Our scorer is the Consistency Judge introduced above (validated at $91.5\%$ human agreement); to reduce single-model bias we apply it as an \emph{ensemble}, defining $J_{\text{con}}(q,y,y^\star)\in[0,1]$ as the mean consistency score between a candidate response $y$ and a reference $y^\star$ for query $q$, averaged over four LLMs (DeepSeek-R1~\cite{Guo-GRPO}, DeepSeek-V3.2~\cite{liu2025deepseekv32}, Qwen3-Max~\cite{qwen3max}, and QwQ-Plus~\cite{alibabacloud2026qwqplus}). Scoring against the single logged reference $y^\star_{0}$ yields the \textbf{Single-ECS} (Single-GT Ensemble Consistency Score), $S_{\text{Single-ECS}}(q,y)=J_{\text{con}}(q,y,y^\star_{0})$, whereas taking the best match over the expanded set $\mathcal{Y}^\star(q)$ yields the \textbf{Multi-ECS} (Multi-GT Ensemble Consistency Score):
\begin{equation}
S_{\text{Multi-ECS}}(q,y)=\max_{y^\star\in \mathcal{Y}^\star(q)} \; J_{\text{con}}(q,y,y^\star).
\label{eq:multigt_eval}
\end{equation}
We use Multi-ECS as the primary metric and Single-ECS to gauge alignment with the single logged reference.

\subsection{SFT Stage: Latent Logic Augmentation}
\label{sec:planning_sft}

Standard SFT on expert trajectories often leads to myopic imitation and performance degradation in complex technical-service tasks. To instill latent decision reasoning, we augment training data through two methods. We model the interaction as a Markov Decision Process whose state is the dialogue context up to the current customer query $q_t$, whose action $a_t$ is the agent's response (a tool call or a textual reply), whose transition produces the environment's next message $q_{t+1}$ (from the customer or a tool), and whose reward scores the consistency of $a_t$. We then augment the training data to make these latent decision processes visible.

\textbf{Reasoning Augmentation.}
To teach immediate reasoning, we turn raw single-step state-action pairs into a \emph{decision} SFT dataset $\mathcal{D}_{\text{RA}}$. For each agent response $a_t$ to query $q_t$, GPT-5.2 generates a ``backward'' chain-of-thought rationale $c_t$ justifying the action, and the model is trained to predict the rationale and then the response:
\begin{equation*}
\mathcal{L}_{\text{RA}} = - \mathbb{E}_{(q_t, c_t, a_t) \sim \mathcal{D}_{\text{RA}}} \left[ \log p_\theta(c_t, a_t \mid q_t) \right].
\end{equation*}
This objective forces the model not just to mimic the action $a_t$, but to first generate the underlying thought process $c_t$, thereby internalizing the decision logic (see prompt in Appendix~\ref{app:prompt_reasoning}).

\textbf{Planning-Aware Trajectory Modeling.}
To equip the agent with foresight, we construct Planning-Aware trajectories for short future interactions. We extract three-step future sequences $(a_t, q_{t+1}, a_{t+1})$ following the current customer query $q_t$ in the tickets, representing an agent response, corresponding environment response, and corresponding next agent response. GPT-5.2 rewrites this into a structured planning form $y_t^{\text{PATM}} = (q_t, \tilde{a}_t, \tilde{q}_{t+1}, \tilde{a}_{t+1})$, where $\tilde{a}_t$ is the predicted response to query $q_t$, $\tilde{q}_{t+1}$ is the predicted response of the environment (customer or tool) to $\tilde{a}_t$, and $\tilde{a}_{t+1}$ is the next predicted response to $\tilde{q}_{t+1}$. The learning objective autoregressively generates this trace:
\begin{equation*}
\resizebox{0.98\columnwidth}{!}{$\displaystyle
\mathcal{L}_{\text{PATM}} = - \mathbb{E}_{y_t^{\text{PATM}} \sim \mathcal{D}_{\text{PATM}}} \left[ \log p_\theta(\tilde{a}_t, \tilde{q}_{t+1}, \tilde{a}_{t+1} \mid q_t) \right]
$}.
\end{equation*}
These trajectories form the \emph{planning} SFT set $\mathcal{D}_{\text{PATM}}$, and predicting them lets the model internalize the environment's transition dynamics, moving beyond simple pattern matching. Because reasoning and planning are complementary, we also apply RA to the planning data, prepending the rationale $c_t$ to obtain the reasoning-augmented trace $y_t^{\text{PATM+R}} = (q_t, c_t, \tilde{a}_t, \tilde{q}_{t+1}, \tilde{a}_{t+1})$; in practice the decision set $\mathcal{D}_{\text{RA}}$ and the planning set $\mathcal{D}_{\text{PATM}}$ are mixed for SFT. This combined trace decomposes into four coupled sub-objectives: (i) reasoning $\pi_\theta(c_t \mid q_t)$, (ii) policy execution $\pi_\theta(\tilde{a}_t \mid q_t, c_t)$, (iii) implicit world modeling $\mathcal{P}_\theta(\tilde{q}_{t+1} \mid q_t, c_t, \tilde{a}_t)$, and (iv) contingency planning $\pi_\theta(\tilde{a}_{t+1} \mid q_t, c_t, \tilde{a}_t, \tilde{q}_{t+1})$. Sub-objective (iii) is the core signal for capturing the stochastic environment dynamics (see prompts in Appendix~\ref{app:prompt_planning} and~\ref{app:prompt_rewrite}).

\subsection{RL Stage: Hybrid Reward Modeling}
\label{sec:hybrid_rl}

To refine the policy $\pi_{\text{SFT}}$ obtained from SFT alignment, we address the challenge of reward sparsity and computational cost in reinforcement learning. During interactions, the agent primarily executes either tool invocations (\texttt{call\_tool}) or textual replies (\texttt{reply}). Any action-type mismatch strictly yields a zero reward. For matched actions, while tool calls are deterministically scored via exact parameter matching, assessing the semantic consistency of free-form textual replies is highly subjective. Using powerful LLMs as judges for every RL rollout incurs significant latency. To optimize the trade-off between efficiency and fidelity, we propose an HRM that fuses a lightweight Reranker with an LLM-based Judge.

\textbf{Components: Reranker and Judge.}
(1) \textbf{Reranker ($S_R$):} Rather than relying on dense embeddings, we use an instruction-augmented Qwen3-4B \emph{consistency} Reranker (non-thinking). It serves as a computationally cheap discriminator ($S_R$) that captures logical contradictions often missed by vector similarity (see Appendix~\ref{app:reranker_justification} for a detailed comparison with embedding-based methods).
(2) \textbf{LLM-based Judge ($S_J$):} A larger Qwen3-32B model (thinking-enabled) provides expert-level consistency scores, using the same prompt template as the Consistency Judge in the Dual-Filtering process. To stabilize training, we use a soft-score derived from token probabilities: $\text{Score} = P(\texttt{``Consistent''}) + 0.5 \cdot P(\texttt{``Partially''})$.

\textbf{Single-Interval Cascade Strategy.}
Running the Qwen3-32B Judge for every RL rollout is prohibitively expensive, incurring approximately 10$\times$ the inference latency compared with the lightweight Reranker. To mitigate this, we employ a cascade strategy $R_\theta(S_R, S_J)$ that approximates the oracle reward by combining the Reranker score $S_R$ and the Judge score $S_J$. Within a ``trust interval'' $[\tau_a,\tau_b]$ the cheap Reranker is trusted directly (Fast Pass); outside it, the two scores are blended with mixing weights $w_1,w_2\in[0,1]$:
\begin{equation}
\resizebox{\columnwidth}{!}{$\displaystyle
R_{\theta}(S_R,S_J)=
\begin{cases}
w_1 S_R + (1-w_1) S_J, & S_R < \tau_a \;\,\text{(Mix)}\\
S_R, & \tau_a \le S_R \le \tau_b \;\,\text{(Fast Pass)}\\
w_2 S_R + (1-w_2) S_J, & S_R > \tau_b \;\,\text{(Mix)}
\end{cases}
$}
\label{eq:single_interval_cascade_main}
\end{equation}
The cascade parameters $\theta=\{\tau_a,\tau_b,w_1,w_2\}$ are fitted once on a held-out set by maximizing the Spearman rank correlation between the cascade output $R_\theta(S_R,S_J)$ and the ensemble Consistency-Judge teacher score; the formal objective and the fitted values are given in Appendix~\ref{app:hyperparameters}.

By routing clear-cut samples (Fast Pass) to the fast Reranker and reserving the costly Judge for ambiguous cases, the HRM reduces the overall reward computation time while maintaining alignment fidelity.

\section{Experiments}
\label{sec:exp}
We evaluate the framework on a Qwen3-4B agent \cite{yang2025qwen3}: after the setup (Sec.~\ref{sec:exp_setup}), we ablate the SFT and RL components---RA, PATM, the Hybrid Reward, and Multi-GT---on offline ECS (Sec.~\ref{sec:exp_components}), compare with matched optimization and routing baselines and test transfer across backbones (Sec.~\ref{sec:exp_generality}), analyze reward calibration and ranking fidelity (Sec.~\ref{sec:exp_calibration}), and report a paired production study of real adoption (Sec.~\ref{sec:exp_online}).

\subsection{Experimental Setup}
\label{sec:exp_setup}

\textit{Dataset \& Metrics.}
Cloud platforms must resolve large volumes of technical-service tickets in which engineers diagnose issues, invoke diagnostic tools, and advise customers under tight correctness and latency constraints. We draw a proprietary technical-service dataset from this setting, with 10k queries each for the decision (RA) and planning (PATM) SFT sets (Sec.~\ref{sec:planning_sft}), alongside 5120/1k/1k queries for RL training/validation/test. To prevent leakage, complete tickets are split temporally before deriving RA, PATM, or Multi-GT examples; all turns from a ticket stay together, the SFT tickets are disjoint from the RL splits, and the preliminary policy uses only training tickets. PATM future turns serve only as training targets. Using our dual-filtering pipeline (Sec.~\ref{sec:multi_gt_foundation}), we expand the single logged reference into a \texttt{Multi-GT} dataset across all splits.
We report \textbf{Multi-ECS} (primary) and \textbf{Single-ECS}, both defined in Sec.~\ref{sec:multi_gt_foundation}.

\textit{Training Setup \& Baselines.}
For SFT strategies, we ablate the two latent-logic components---Reasoning Augmentation (RA) and Planning-Aware Trajectory Modeling (PATM)---giving the additive variants \texttt{SFT}, \texttt{SFT+RA}, \texttt{SFT+PATM}, and \texttt{SFT+RA+PATM}.
In the RL stage, we employ the DAPO algorithm \cite{yu2025dapo}. For reward signals, we compare: (1) \texttt{Reranker}, (2) \texttt{Hard Judge}, (3) \texttt{Soft Judge}, and (4) \textbf{\texttt{Hybrid Reward}}. For data construction, we compare \texttt{Single-GT} RL (reward: the cascade score against the single logged reference) with our \textbf{\texttt{Multi-GT}} expansion (reward: the \emph{maximum} cascade score over the validated set $\mathcal{Y}^\star(q)$). We additionally adapt DPO~\cite{rafailov2023direct} to the Multi-GT data (Multi-GT DPO) and implement RACER adaptive judge routing~\cite{zhang2026reasoningnotfree} with the same Qwen3-4B backbone, data, Judge service, and evaluation protocol. To ensure a fair comparison, all RL models are evaluated at the same fixed checkpoint after 20 episodes. Detailed configurations are provided in Appendix~\ref{app:hyperparameters} and~\ref{app:matched_baselines}.

\subsection{Main Results}
\label{sec:exp_components}

We isolate and validate the proposed components below, summarizing SFT phase results in \textbf{Table~\ref{tab:sft_analysis_4b}} and RL phase results in \textbf{Table~\ref{tab:rl_analysis_4b}}.

\textbf{RQ1: Impact of SFT Strategies and Latent Logic.}
\label{sec:exp_sft}
Table~\ref{tab:sft_analysis_4b} shows that plain \texttt{SFT} slightly degrades Multi-ECS, highlighting ``myopic'' imitation. Adding backward-looking reasoning (\texttt{SFT+RA}) mitigates this, and further adding forward-looking planning (\texttt{SFT+RA+PATM}) achieves peak Multi-ECS (0.337) and doubles Call Tool Accuracy ($0.139 \to 0.279$). These results demonstrate the significance of latent decision logic in complex technical-service tasks. We default to \texttt{SFT+RA+PATM} for subsequent RL experiments.

\begin{table}[tbp]
    \centering
    \small
    \caption{SFT strategy analysis (RQ1). RA and PATM each improve Multi-ECS and tool-use accuracy; their combination is best.}
    \label{tab:sft_analysis_4b}
    \resizebox{\columnwidth}{!}{%
    \begin{tabular}{lccc}
        \toprule
        \textbf{Method} & \textbf{Multi-ECS} & \textbf{Single-ECS} & \textbf{Call Tool Acc} \\
        \midrule
        Untrained Qwen3-4B & 0.299 & 0.178 & 0.082 \\
        \midrule
        SFT          & 0.293 & 0.193 & 0.096 \\
        SFT+RA       & 0.319 & 0.224 & 0.139 \\
        SFT+PATM     & 0.326 & 0.233 & 0.149 \\
        \textbf{SFT+RA+PATM} & \textbf{0.337} & \textbf{0.242} & \textbf{0.279} \\
        \bottomrule
    \end{tabular}}
\end{table}

\textbf{RQ2: Impact of SFT Base and Reward Design on RL.}
\label{sec:exp_reward}
Table~\ref{tab:rl_analysis_4b} starts from our default (Ours: \texttt{SFT+RA+PATM+Hybrid+Single-GT}, 0.429 Multi-ECS) and ablates one component at a time. \emph{Initialization} is decisive: skipping \texttt{SFT} (\texttt{non-SFT}) collapses alignment to 0.348, dropping either latent-logic component caps it at 0.406--0.407, and plain \texttt{SFT} reaches only 0.353---confirming that Latent Logic Augmentation gives RL an indispensable starting point. \emph{Reward}: the Reranker alone underperforms (0.389) due to lexical reliance, the Hard/Soft Judges reach 0.406/0.413, while our Hybrid Reward attains 0.429 \emph{and}, via the cascade, cuts reward computation from 170 to 129 GPU-hours (a 24\% reduction).

\textbf{RQ3: Effectiveness of Multi-GT Expansion.}
\label{sec:exp_multigt}
Upgrading to the \texttt{Multi-GT} dataset (Block 3) raises Multi-ECS to \textbf{0.441}; the slight Single-ECS drop ($0.357\to0.347$) reflects that \texttt{Single-GT} forces the policy toward one arbitrary phrasing, whereas \texttt{Multi-GT} rewards semantically diverse yet valid resolutions and mitigates output homogenization. Consistently, \texttt{Multi-GT} sustains higher policy entropy while \texttt{Single-GT} collapses (Appendix~\ref{app:entropy}). To test whether this gain depends on reference cleaning, we evaluate the outputs of every configuration against both the unfiltered candidate pool and the final human-validated pool. In both evaluations, the Full model remains ahead of Hybrid+Single-GT in Multi-ECS, with detailed audit statistics and complete raw-vs.-cleaned results provided in Appendix~\ref{app:multigt_stats}.

A per-category breakdown (Appendix~\ref{app:per_category}) shows the Hybrid's gains concentrate on the ambiguous \texttt{inquire}/\texttt{clarify} actions and that it attains the highest action-type accuracy; three case studies (Appendix~\ref{app:case_studies}) illustrate PATM's planning, the cascade's resistance to reward hacking, and Multi-GT's useful diversity.

\begin{table}[tbp]
    \centering
    \small
    \caption{RL Adaptation and Robust Noise Reduction Analysis (RQ2--3). We establish a default configuration (\textbf{Ours}) and independently ablate its components. Upgrading the data to \texttt{Multi-GT} yields the final peak performance. Paired one-sided $t$-tests on Multi-ECS confirm \textbf{Ours} significantly exceeds the Soft-Judge reward ($p<0.05$), the Reranker-only reward, and the \texttt{plain-SFT} initialization (both $p<0.001$).}
    \label{tab:rl_analysis_4b}
    \resizebox{\columnwidth}{!}{%
    \begin{tabular}{lcc}
        \toprule
        \textbf{Configuration Variant} & \textbf{Multi-ECS} & \textbf{Single-ECS} \\
        \midrule
        \textbf{Ours (SFT+RA+PATM + Hybrid + Single-GT)} & 0.429 & \textbf{0.357} \\
        \midrule
        \multicolumn{3}{l}{\textit{\textbf{(1) Varying SFT Initialization}}} \\
        \quad $\rightarrow$ replace with non-SFT          & 0.348 & 0.231 \\
        \quad $\rightarrow$ replace with \texttt{SFT}     & 0.353 & 0.331 \\
        \quad $\rightarrow$ replace with \texttt{SFT+RA}  & 0.406 & 0.336 \\
        \quad $\rightarrow$ replace with \texttt{SFT+PATM} & 0.407 & 0.346 \\
        \midrule
        \multicolumn{3}{l}{\textit{\textbf{(2) Varying Reward Signal}}} \\
        \quad $\rightarrow$ replace with Reranker Only          & 0.389 & 0.309 \\
        \quad $\rightarrow$ replace with Hard Judge             & 0.406 & 0.336 \\
        \quad $\rightarrow$ replace with Soft Judge             & 0.413 & 0.344 \\
        \midrule
        \multicolumn{3}{l}{\textit{\textbf{(3) Varying Training Data Construction}}} \\
        \quad $\rightarrow$ \textbf{upgrade to Multi-GT (Full Framework)} & \textbf{0.441} & 0.347 \\
        \bottomrule
    \end{tabular}}
\end{table}

\subsection{Generality and Matched Baselines}
\label{sec:exp_generality}

Our generality evaluation spans prompting-only comparisons, backbone transfer, and matched training baselines (Appendix~\ref{app:matched_baselines}, Table~\ref{tab:external_results}). Under matched training configurations, Llama-3.1-8B and Qwen3-32B preserve the progression from SFT to Reranker-only RL and then Hybrid-reward RL, with Multi-GT producing the highest Multi-ECS on both backbones. Under the matched Qwen3-4B Multi-GT setup, our method reaches 0.347/0.441 Single-/Multi-ECS, compared with 0.291/0.373 for Multi-GT DPO and 0.314/0.421 for RACER, while using fewer Judge calls and reward GPU-hours than RACER. Taken together, the consistent transfer trend and matched-baseline advantage show that the framework's gains extend beyond a single backbone or post-training objective.

\subsection{Calibration and Ranking Fidelity}
\label{sec:exp_calibration}

\begin{figure}[tbp]
    \centering
    \includegraphics[width=\linewidth]{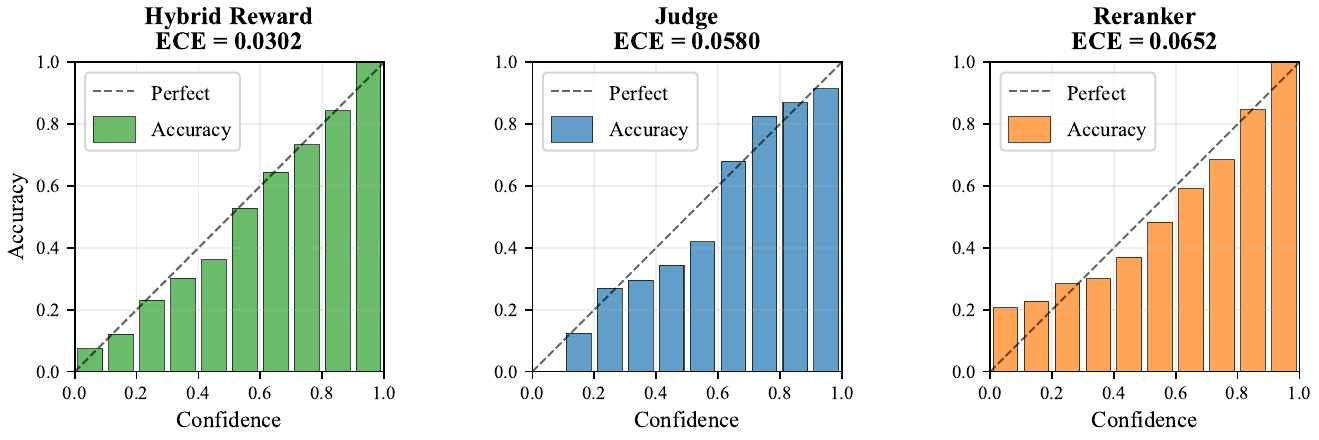}
    \caption{Expected Calibration Error (ECE) reliability diagrams for the three reward configurations. Bars show observed accuracy against predicted confidence; the dashed line denotes perfect calibration.}
    \label{fig:calibration}
\end{figure}

\textbf{Calibration.}
Figure~\ref{fig:calibration} shows the Hybrid reward is markedly better calibrated (ECE\,=\,0.0302) than the standalone Judge (0.0580) and Reranker (0.0652): the cascade's region-gating offsets the two signals' \emph{opposite-signed} biases---in the low-confidence Mix zone the Reranker is under-confident ($-0.012$) and the Judge over-confident ($+0.064$), while in the Fast-Pass zone the Reranker is the better calibrated of the two ($0.041$ vs.\ $0.047$ mean absolute gap), so routing clear-cut cases to it even improves calibration.

\textbf{Ranking fidelity.} In ranking terms (Spearman correlation with the oracle $\rho$), the Reranker alone reaches only $\rho=0.3295$ while the Hybrid ($\rho=0.5172$) marginally surpasses the Judge ($\rho=0.5166$), even when many samples are offloaded to the cheap Reranker, resulting in fidelity at much lower cost (see Appendix~\ref{app:ranking}, Figure~\ref{fig:spearman}).

\subsection{Online Adoption Rate}
\label{sec:exp_online}

To assess whether offline gains transfer to production, we conduct a paired, blinded retrospective study on 1{,}000 randomly sampled production tickets and report the \emph{adoption rate}---the share of model replies that engineers accept for use (Table~\ref{tab:online_adoption}). The full framework reaches the highest adoption rate (46.63\%) and significantly outperforms the untrained Qwen3-4B, SFT+RA+PATM, the prompted Qwen3.7-Max~\cite{qwen37}, Soft Judge, and Reranker-only under two-sided McNemar tests ($p=1.6\!\times\!10^{-14}$, $9.9\!\times\!10^{-10}$, $1.9\!\times\!10^{-5}$, $0.042$, and $0.030$, respectively). Mirroring Table~\ref{tab:rl_analysis_4b}, replacing the Hybrid reward with either single scorer or dropping Multi-GT lowers adoption. This suggests that task-specific adaptation is important in this Cloud technical-service setting.

\begin{table}[tbp]
    \centering
    \small
    \renewcommand{\arraystretch}{0.95}
    \caption{Online adoption rate (\%) from a retrospective production study. The RL-adaptation rows mirror the one-component ablations of \textbf{Ours} in Table~\ref{tab:rl_analysis_4b}.}
    \label{tab:online_adoption}
    \resizebox{\columnwidth}{!}{%
        \begin{tabular}{lc}
        \toprule
        \textbf{Model / Configuration} & \textbf{Adoption (\%)} \\
        \midrule
        \multicolumn{2}{l}{\textit{\textbf{Untrained baselines}}} \\
        \quad Qwen3-4B                    & 32.80 \\
        \quad Qwen3.7-Max                 & 39.14 \\
        \addlinespace[2pt]
        \multicolumn{2}{l}{\textit{\textbf{SFT only}}} \\
        \quad SFT+RA+PATM                 & 35.51 \\
        \addlinespace[2pt]
        \multicolumn{2}{l}{\textit{\textbf{RL adaptation (reward ablations)}}} \\
        \quad reward $\to$ Reranker Only  & 44.77 \\
        \quad reward $\to$ Soft Judge     & 44.50 \\
        \addlinespace[2pt]
        Ours (SFT+RA+PATM + Hybrid)       & 46.01 \\
        \textbf{Ours + Multi-GT (Full)}   & \textbf{46.63} \\
        \bottomrule
    \end{tabular}
    }
\end{table}

\section{Conclusion}
\label{sec:conclusion}
We presented a framework for technical-service LLM agents that couples latent-logic SFT, Multi-GT supervision, and Hybrid Reward RL. On real Cloud technical-service tasks, it improves Multi-ECS, reduces reward computation by 24\%, and yields the highest online adoption rate. Matched Multi-GT DPO and RACER comparisons support its advantage over relevant alternatives, while transfer experiments reproduce the gains on Llama-3.1-8B and Qwen3-32B.
Together, latent-logic supervision, diversity-aware Multi-GT training, and cost-efficient Hybrid Reward Modeling make compact service agents more accurate, efficient, and deployable.

\section*{Limitations}
Our study has several limitations.
First, all experiments use a proprietary, real-world Cloud technical-service dataset. The raw tickets cannot be released because they contain customer, infrastructure, and incident information, limiting direct replication on the same corpus. We instead document the construction pipeline, prompts, optimization settings, and evaluation protocol so that the method can be reimplemented on other technical-service corpora.
Second, both latent-logic augmentation (RA and PATM) and Multi-GT expansion rely on large teacher models for rationale/trajectory generation and judging; reducing this dependence on costly teacher models remains an open direction.
Third, the Hybrid Reward cascade is calibrated offline and its interval and mixing weights remain fixed during RL, which may be suboptimal as the policy distribution shifts; adaptive or online-updated reward routing and asynchronous judge updates are left to future work.
Finally, our evaluation centers on Cloud technical-service tickets; generalizing the framework to other high-stakes domains characterized by implicit logic and response diversity (e.g., medical or legal assistance) remains to be validated.

\section*{Ethical Considerations}
The proprietary technical-service tickets used in this study may contain sensitive customer and infrastructure information. We therefore do not release raw records and de-identify all examples reported in the paper. Because incorrect operational advice or premature tool use could affect production systems, every generated reply is reviewed and approved by an engineer before it can reach a customer.

\section*{Acknowledgments}
This work was supported by Alibaba Group through Alibaba Research Intern Program.
\par\noindent Additional support was provided by the Science \& Technology Commission of Shanghai Municipality (No.~23JC1400800); Ji Hua Laboratory S\&T Program (No.~X250881UG250); the Fudan University--CIOMP Joint Fund and the key projects of the ``Double First-Class'' initiative of Fudan University; and the Lingang Laboratory (Grant No.~LGL-1987).

\bibliography{references}

\clearpage
\appendix
\section{Dataset Construction Statistics}
\label{app:multigt_stats}

We detail the composition of the Multi-GT dataset and the provenance of the added references. Table~\ref{tab:multigt_stats_grouped} reports, for each split, the number of queries, the Single-GT and Multi-GT reference counts, and the breakdown of newly added references by source---consistency-judge-approved online rollouts and offline candidates, together with utility-judge-approved alternatives. Across all splits, Multi-GT expansion roughly doubles the number of valid references, with the largest contribution coming from consistency-judge-approved online rollouts.

\begin{strip}
\centering
\small
\setlength{\tabcolsep}{5pt}
\resizebox{\textwidth}{!}{%
\begin{tabular}{lrrrr|rrrr}
\toprule
& \multicolumn{4}{c|}{\textbf{Dataset Size}} & \multicolumn{4}{c}{\textbf{Multi-GT Expansion Breakdown}} \\
\cmidrule(lr){2-5}\cmidrule(lr){6-9}
\textbf{Split} &
\textbf{\#Queries} &
\textbf{Single-GT} &
\textbf{Multi-GT} &
\textbf{+Added} &
\textbf{Con. Judge(online)} &
\textbf{Con. Judge(offline)} &
\textbf{Utility Judge} &
\textbf{Expand \%} \\
\midrule
Test  & 1000  & 1000  & 1975  & 975   & 543  & 34  & 398  & 97.50 \\
Val   & 1000  & 1000  & 2091  & 1091  & 671  & 41  & 379  & 109.10 \\
Train & 5120 & 5120 & 10127 & 5007 & 2834 & 143 & 2030 & 97.79 \\
\bottomrule
\end{tabular}}
\captionof{table}{\textbf{Statistics of Multi-GT construction and expansion sources.}
Left: dataset sizes before/after Multi-GT expansion. Right: breakdown of newly added references by source.
\textbf{+Added} $=$ Multi-GT $-$ Single-GT, and
\textbf{Expand \%} $= \frac{\text{+Added}}{\text{Single-GT}}\times 100$.}
\label{tab:multigt_stats_grouped}
\end{strip}

\paragraph{Human Reference Audit and Raw-vs.-Cleaned Robustness.}

Every test reference added by either construction branch was checked by humans before the submitted scores were computed. The audit evaluated 614 Consistency candidates and rejected 37 (precision $577/614=93.97\%$); it evaluated 474 Utility candidates and rejected 76 (precision $398/474=83.97\%$). Together with the 1,000 logged references, the retained 577 Consistency and 398 Utility additions produce the 1,975-reference test set in Table~\ref{tab:multigt_stats_grouped}. Utility validation is harder because a useful answer may follow a different resolution path from the logged action, whereas Consistency candidates can be compared directly with an existing expert response.

\newpage
Table~\ref{tab:raw_cleaned_audit} rescored every configuration against the pre-check (raw) candidate sets. Cleaning changes the Consistency-only scores by at most 0.003 (0.001 on average) and the Utility-only scores by at most 0.015. All non-tied orderings are preserved; the only apparent swap is between two cleaned configurations tied at 0.406 after reporting precision (SFT+RA initialization and Hard Judge). Most importantly, Full exceeds Hybrid+Single-GT by 0.008 on the raw set and by 0.012 on the cleaned set. Human filtering therefore does not manufacture the Multi-GT gain.

\clearpage
\begin{strip}
\centering
\scriptsize
\setlength{\tabcolsep}{3pt}
\captionof{table}{Reference-audit robustness. ``Clean'' uses human-validated references; ``Raw'' retains the pre-check candidates. +Cons. and +Util. score against the logged reference plus additions from the indicated branch. Bold marks the best value in each column.}
\label{tab:raw_cleaned_audit}
\resizebox{\textwidth}{!}{%
\begin{tabular}{lccccccc}
\toprule
\textbf{Configuration} & \textbf{Multi clean} & \textbf{Multi raw} & \textbf{Single} & \textbf{+Cons. clean} & \textbf{+Cons. raw} & \textbf{+Util. clean} & \textbf{+Util. raw} \\
\midrule
Untrained Qwen3-4B & 0.299 & 0.315 & 0.178 & 0.210 & 0.213 & 0.282 & 0.294 \\
SFT & 0.293 & 0.303 & 0.193 & 0.218 & 0.220 & 0.272 & 0.279 \\
SFT+RA & 0.319 & 0.326 & 0.224 & 0.250 & 0.253 & 0.298 & 0.302 \\
SFT+PATM & 0.326 & 0.335 & 0.233 & 0.262 & 0.264 & 0.307 & 0.315 \\
SFT+RA+PATM & 0.337 & 0.350 & 0.242 & 0.270 & 0.271 & 0.319 & 0.330 \\
Hybrid + Single-GT & 0.429 & 0.442 & \textbf{0.357} & \textbf{0.375} & \textbf{0.376} & 0.417 & 0.429 \\
RL init: non-SFT & 0.348 & 0.356 & 0.231 & 0.264 & 0.265 & 0.330 & 0.336 \\
RL init: SFT & 0.353 & 0.363 & 0.331 & 0.336 & 0.337 & 0.348 & 0.362 \\
RL init: SFT+RA & 0.406 & 0.413 & 0.336 & 0.356 & 0.356 & 0.395 & 0.401 \\
RL init: SFT+PATM & 0.407 & 0.420 & 0.346 & 0.362 & 0.363 & 0.399 & 0.414 \\
Reward: Reranker only & 0.389 & 0.396 & 0.309 & 0.328 & 0.329 & 0.376 & 0.382 \\
Reward: Hard Judge & 0.406 & 0.414 & 0.336 & 0.352 & 0.354 & 0.398 & 0.406 \\
Reward: Soft Judge & 0.413 & 0.422 & 0.344 & 0.363 & 0.365 & 0.404 & 0.412 \\
\textbf{Full (Hybrid + Multi-GT)} & \textbf{0.441} & \textbf{0.450} & 0.347 & 0.367 & 0.368 & \textbf{0.430} & \textbf{0.436} \\
\bottomrule
\end{tabular}}
\end{strip}

\section{Consistency Judge Reliability}
\label{app:judge_reliability}
We validate the Consistency Judge against 152 human expert annotations, on which it attains an overall accuracy of $91.45\%$ ($139/152$). Across standard interval estimators the $95\%$ confidence interval stays within the $\pm5\%$ margin commonly accepted in the LLM-evaluation literature (e.g., Wilson $[0.859,0.949]$), and a non-parametric bootstrap ($B=10{,}000$ resamples) agrees ($95\%$ CI $[0.868,0.954]$). To rule out agreement by chance under class imbalance (91 positive vs.\ 61 negative samples), we report Cohen's $\kappa=0.82$, i.e.\ ``almost perfect agreement'' on the Landis--Koch scale. The sample is also large enough: attaining a $95\%$ confidence interval within the $\pm5\%$ margin at this accuracy requires at least $140$ annotations under the most conservative (Clopper--Pearson) estimator, which our $n=152$ exceeds. This validation scale exceeds the 80-example expert-agreement study reported for MT-Bench~\citep{Zheng-judging}.

\section{Qualitative Case Studies}
\label{app:case_studies}

We present three representative case studies that illustrate, at the level of individual rollouts, how each component changes model behavior: for PATM we contrast reasoning traces, for Hybrid Reward we expose the underlying reward signals, and for Multi-GT we compare sampled responses against an expanded reference set.

\begin{table*}[t]
\centering
\small
\caption{Case study for PATM: a \texttt{LATERAL} query succeeds on test data in the query IDE but fails on the real table in the data platform. Abbreviated reasoning traces (\texttt{<think>}) contrast the two decision processes; the score is the Multi-ECS against the reference set.}
\label{tab:case_patm}
\begin{tabular}{p{0.12\textwidth}p{0.81\textwidth}}
\toprule
\textbf{Component} & \textbf{Content} \\
\midrule
Context & The customer uses \texttt{LATERAL} to expand a JSON array. The test code runs fine in the query IDE, but after switching to the real table it fails in the data platform with ``Feature not supported: LATERAL''. \\
\midrule
Ground Truth & (\texttt{inquire}) ``@user~Please connect the real table and run it in the query IDE, then send me the full execution log; I have not found the corresponding error record yet.'' \\
 & \textit{(Expert strategy: re-run the diagnosis in a clean query-IDE environment and collect full logs)} \\
\midrule
Baseline \texttt{<think>} & ``No tools are available right now\dots{} the help doc says \texttt{LATERAL} is not supported\dots{} I should advise avoiding this syntax.'' \\
 & \textit{(Leaps from ``no tools available'' to a generic workaround, ignoring the environment discrepancy)} \\
\cmidrule{2-2}
Baseline Response & ``@user~Your current version is 3.0, which does not support \texttt{LATERAL}; I recommend using another method to process the JSON array.'' \\
 & (\ding{55}~Premature \texttt{solution}; Multi-ECS $=0.00$) \\
\midrule
PATM \texttt{<think>} & ``The query runs in the data platform but the real table errors out, so I must confirm the difference in execution environment; different query-IDE versions may support different features; no tool can fetch the environment automatically, so I can only ask where it was run---hence the next step should be \texttt{inquire}.'' \\
 & \textit{(Spots the environment-dependent contradiction $\rightarrow$ inventories the missing information $\rightarrow$ plans a targeted diagnostic action)} \\
\cmidrule{2-2}
PATM Response & ``@user~Are you running this SQL in the query IDE? Please connect the real table in the query IDE, run it, and send me the full execution log---I can't find the corresponding error record on my side yet.'' \\
 & (\ding{51}~\texttt{inquire} that mirrors the expert strategy; Multi-ECS $=\mathbf{1.00}$) \\
\bottomrule
\end{tabular}
\end{table*}

\textbf{PATM Case Analysis.} This case exemplifies the ``myopic imitation'' problem of standard SFT. The reasoning traces expose the difference: the baseline's \texttt{<think>} jumps from ``no available tools'' to a documentation-based workaround, overlooking the decisive clue---the same query succeeds in the query IDE but fails on the real table, which points to an environment-specific issue rather than a syntax limitation. The PATM model instead verbalizes a short plan---it identifies the contradiction, notes that no tool can recover the execution environment automatically, and therefore decides to \texttt{inquire}---exactly the trajectory $(q_t,\tilde{a}_t,\tilde{q}_{t+1},\tilde{a}_{t+1})$ it was trained to generate. The perfect score is credible because the Consistency Judge rewards \emph{strategic} alignment rather than surface form: both the PATM reply and the reference adopt the identical resolution path---defer judgement, ask the user to re-run against the real table in the query IDE, and return the full logs---whereas the baseline diverges into a premature solution and scores $0.00$.

\begin{table*}[t]
\centering
\small
\caption{Case study for Hybrid Reward: the customer sees several ``instance VIP migration'' events in the managed-database console and asks whether the instance will be migrated away---\emph{without providing an instance ID}. The reference strategy is to \texttt{inquire} for the instance ID before diagnosing; the two wrong answers are \emph{plausible but premature} solutions. We list the responses of three policies, each trained with a different reward signal.}
\label{tab:case_hybrid}
\begin{tabular}{p{0.15\textwidth}p{0.79\textwidth}}
\toprule
\textbf{Component} & \textbf{Content} \\
\midrule
Context & In the managed-database event center the customer sees several ``instance VIP migration'' events and asks whether this means the instance will be migrated away. No instance ID is provided in the conversation. \\
\midrule
Ground Truth & (\texttt{inquire}) 
``@user~To check whether this instance VIP migration will affect your instance, could you please send the instance ID?''\\
 & \textit{(Expert strategy: obtain the instance ID before diagnosing the event)} \\
\midrule
$r_R$: Reranker-trained & ``@user~Yes, this event is a migration of the instance's virtual IP address; it is a normal maintenance operation and will not affect instance availability.'' \\
 & (\ding{55}~confident but premature \texttt{solution} from generic experience) \\
\midrule
$r_J$: Judge-trained & ``@user~An instance VIP migration usually means the virtual IP configuration has changed, for network optimization or maintenance; it is normal operational behavior and does not cause the instance itself to be migrated away.'' \\
 & (\ding{55}~elaborate, authoritative---but still a premature \texttt{solution}) \\
\midrule
$r_H$: Hybrid-trained & ``@user~For the instance VIP migration event you mentioned, could you provide a specific instance ID or event ID?'' \\
 & (\ding{51}~\texttt{inquire} that matches the expert strategy) \\
\bottomrule
\end{tabular}
\end{table*}

\begin{table}[t]
\centering
\small
\caption{Reward scores assigned to the three responses of Table~\ref{tab:case_hybrid} by each signal. $S_R$/$S_J$ are the Reranker/Judge consistency scores; $R_\theta$ is the Hybrid cascade (Eq.~\ref{eq:single_interval_cascade_main}) with $[\tau_a,\tau_b]\!=\![0.6,0.9]$, $w_1\!=\!0.05$, $w_2\!=\!0.72$; Multi-ECS is the oracle. Mix$^{\uparrow}$/Mix$^{\downarrow}$ denote the upper ($S_R\!>\!\tau_b$) and lower ($S_R\!<\!\tau_a$) mixing branches. Bold marks the top-scored response per signal: the Reranker and the Judge each rank a different \emph{wrong} answer highest, whereas the Hybrid and the oracle both rank $r_H$ highest.}
\label{tab:hybrid_reward_signals}
\resizebox{\columnwidth}{!}{%
\begin{tabular}{lccc}
\toprule
 & $r_R$ & $r_J$ & $r_H$ \\
 & (solution) & (solution) & (inquire) \\
\midrule
$S_R$ (Reranker)    & \textbf{0.94} & 0.50 & 0.88 \\
$S_J$ (Judge)       & 0.30 & \textbf{0.82} & 0.60 \\
\midrule
Cascade region      & Mix$^{\uparrow}$ & Mix$^{\downarrow}$ & Fast Pass \\
$R_\theta$ (Hybrid) & 0.76 & 0.80 & \textbf{0.88} \\
\midrule
Multi-ECS (oracle)  & 0.02 & 0.01 & \textbf{1.00} \\
\bottomrule
\end{tabular}}
\end{table}

\textbf{Hybrid Reward Case Analysis.} Because the customer gives no instance ID, the only correct action is to \texttt{inquire} ($r_H$); the two solutions ($r_R$, $r_J$) are fluent and on-topic but premature, and the oracle accordingly scores them near zero. Table~\ref{tab:hybrid_reward_signals} shows that the two single signals fail in \emph{complementary} ways. The Reranker, driven by surface fluency and topical relevance, is most fooled by the confident, generic solution $r_R$, to which it assigns $0.94$---above even the correct inquiry ($0.88$); a policy trained on the Reranker alone is thus pulled toward assertive premature answers (reward hacking). The Judge, conversely, exhibits a verbosity/solution bias: it over-rewards the elaborate, authoritative-sounding solution $r_J$ ($0.82$) and under-rewards the terse clarifying question $r_H$ ($0.60$), so a Judge-only policy is pulled toward over-explaining. Crucially, \emph{each single signal ranks a different wrong answer highest}, so neither alone tracks the oracle---this is precisely why the Judge column does not dominate the Hybrid.

The cascade recovers the correct ordering by routing each response to whichever signal is reliable in its selected score region (Eq.~\ref{eq:single_interval_cascade_main}). The Reranker's over-confident score on $r_R$ falls in the upper tail ($S_R\!>\!\tau_b$), where $28\%$ of the Judge tempers it: $R_\theta(r_R)\!=\!0.72\!\times\!0.94+0.28\!\times\!0.30\!=\!0.76$. The Judge's inflated score on $r_J$ sits in the Reranker's low region ($S_R\!<\!\tau_a$, lower Mix), giving $R_\theta(r_J)\!=\!0.05\!\times\!0.50+0.95\!\times\!0.82\!=\!0.80$. The correct inquiry $r_H$ lands in the Reranker's calibrated mid-range, where the Fast Pass trusts it directly: $R_\theta(r_H)\!=\!0.88$. The Hybrid therefore ranks $r_H$ highest ($0.88\!>\!0.80\!>\!0.76$), matching the oracle, whereas the Reranker would have rewarded $r_R$ and the Judge would have rewarded $r_J$. Accumulated over thousands of rollouts, this complementary gating steers the Hybrid policy toward the correct \texttt{inquire} behavior at a fraction of the Judge's cost.

\begin{table*}[t]
\centering
\small
\caption{Case study for Multi-GT: the customer asks how to identify which tables have already been synchronized in the data platform. The reference set admits one \texttt{clarify} and two \texttt{solution} strategies. We draw four samples from each policy; \textbf{Avg.\ Multi-ECS} is averaged over the four samples.}
\label{tab:case_multigt}
\begin{tabular}{p{0.12\textwidth}p{0.81\textwidth}}
\toprule
\textbf{Component} & \textbf{Content} \\
\midrule
Context & The customer has multiple data sources pulling data from business databases and is unsure which tables have already been pulled; for each new extraction they are also unsure whether it was done before, and ask where this can be identified. \\
\midrule
Ground Truth & \texttt{GT-1} (\texttt{clarify}): ``Are you using an offline synchronization task to pull the data?'' \\
(3 references) & \texttt{GT-2} (\texttt{solution}): ``On the Data Source Management page, check the created data sources and their sync-task records to confirm the synchronized tables.'' \\
 & \texttt{GT-3} (\texttt{solution}): ``Go to Data Development > Project Directory > Offline Sync Nodes to view the configured sync tasks and execution records.'' \\
\midrule
Single-GT & All four samples collapse to the same \texttt{inquire} (workspace/project ID): \\
(Avg.\ Multi-ECS & ``(1) Could you provide the specific workspace name or project ID? (2) Which data source's table specifically? (3) Which workspace are you operating in? (4) Could you provide the workspace name?'' \\
 $=0.110$) & (\ding{55}~Homogenized; covers \emph{none} of the \texttt{solution} references) \\
\midrule
Multi-GT & Three \texttt{inquire}/\texttt{clarify} from different angles $+$ one \texttt{solution}: \\
(Avg.\ Multi-ECS & ``(1) Is it a specific table you need to check? (2) Which data source's table needs to be confirmed as synchronized? (3) Are you using an offline synchronization task to pull the data? (4) I suggest checking the list of created data sources on the Data Source Management page of the data-platform console.'' \\
 $=\mathbf{0.286}$) & (\ding{51}~Response~3 matches \texttt{GT-1}; Response~4 matches \texttt{GT-2}/\texttt{GT-3}) \\
\bottomrule
\end{tabular}
\end{table*}

\textbf{Multi-GT Case Analysis.} This case makes the knowledge-homogenization failure of Single-GT training concrete. Optimized toward the single logged reference (which happens to be an inquiry), the Single-GT policy collapses all four samples into surface rephrasings of one question---asking for the workspace or project ID---so none of them reaches the \texttt{solution} references, and the average Multi-ECS is only $0.110$. The Multi-GT policy instead reproduces the breadth of the reference set: the third sample matches the \texttt{clarify} reference \texttt{GT-1}, and the fourth pivots to a \texttt{solution} that matches \texttt{GT-2}/\texttt{GT-3}, raising the average Multi-ECS to $\mathbf{0.286}$ ($2.6\times$). Crucially, the improvement comes precisely from the references \emph{beyond} the single logged one: because the expanded set rewards the solution path as well as the inquiry path, more of the policy's samples land on a valid reference. This is ``useful diversity''---variation across distinct yet valid resolution strategies that \emph{raises} rather than dilutes consistency---which is exactly the behavior a robust technical-service agent should exhibit on ambiguous queries.

\section{Experimental Settings and Hyperparameters}
\label{app:hyperparameters}

We provide the detailed hyperparameter configurations used in our experiments. Table~\ref{tab:sft_hyperparameters} details the settings for the Supervised Fine-Tuning (SFT) stage, which serves as the cold-start initialization for all RL experiments. Table~\ref{tab:rl_hyperparameters} presents the configurations for the Reinforcement Learning stage using 
Decoupled Clip and Dynamic sAmpling Policy Optimization
(DAPO).

\subsection{Supervised Fine-Tuning (SFT) Configuration}

For SFT, we fine-tune the base models on the mixed dataset. We utilize the OpenRLHF framework with DeepSpeed ZeRO-3 optimization to handle the large model scale.

\begin{table*}[h]
\centering
\small
\renewcommand{\arraystretch}{1.2}
\caption{\textbf{Hyperparameters for Supervised Fine-Tuning (SFT).}}
\label{tab:sft_hyperparameters}
\setlength{\tabcolsep}{12pt}
\begin{tabular}{l|l|l}
\toprule
\textbf{Category} & \textbf{Hyperparameter} & \textbf{Value} \\
\midrule
Data
 & Max Sequence Length & 20000 \\

\midrule
Optimization
 & Base Model & Qwen3-4B \\
 & Global Batch Size & 256 \\
 & Learning Rate & $2\times 10^{-6}$ \\
 & Min Learning Rate & $8\times 10^{-7}$ \\
 & LR Scheduler & Cosine \\
 & Warmup Ratio & 0.01 \\
 & Weight Decay & 0.0001 \\
 & Epochs & 1 \\
 & Optimizer & AdamW (ZeRO-3) \\
\bottomrule
\end{tabular}
\end{table*}

\subsection{Reinforcement Learning (RL) Configuration}

\noindent\textbf{HRM threshold fitting.} The Hybrid cascade parameters $\theta=(\tau_a,\tau_b,w_1,w_2)$ (Eq.~\ref{eq:single_interval_cascade_main}) are fitted once on a held-out set, \emph{before} RL and then kept fixed. For each instance $x=(q,y_a,y_b)$ with Reranker score $S_R(x)\in[0,1]$, Judge score $S_J(x)\in[0,1]$, and ensemble teacher score $Y(x)\in[0,1]$ (the Ensemble-Consistency Score), we select $\theta^*$ from the feasible set $\Theta=\{(\tau_a,\tau_b,w_1,w_2)\mid \tau_a,\tau_b,w_1,w_2\in[0,1],\; \tau_b-\tau_a=A\}$, where $A$ is a preset trust-interval width, to maximize the Spearman rank correlation between the cascade output and the teacher over the fitting set $\mathcal{S}=\{x_i\}_{i=1}^N$:
\begin{equation}
\resizebox{\columnwidth}{!}{$\displaystyle
\theta^*
= \arg\max_{\theta\in\Theta}\;
\rho_{\text{spearman}}\!\Big(
\{R_\theta(S_R(x_i),S_J(x_i))\}_{i=1}^N,
\{Y(x_i)\}_{i=1}^N
\Big).
$}
\label{eq:hrm_spearman_obj}
\end{equation}
The fitted configuration is $[\tau_a,\tau_b]=[0.6,0.9]$, $w_1=0.05$, $w_2=0.72$ (Table~\ref{tab:rl_hyperparameters}).

For the RL stage, we employ the DAPO algorithm. We use the Hybrid Reward Mixer as the primary signal.
\begin{table*}[h]
\centering
\small
\renewcommand{\arraystretch}{1.2}
\caption{\textbf{Hyperparameters for Reinforcement Learning (DAPO).}}
\label{tab:rl_hyperparameters}
\setlength{\tabcolsep}{10pt}
\begin{tabular}{l|l|l}
\toprule
\textbf{Category} & \textbf{Hyperparameter} & \textbf{Value} \\
\midrule
Actor Model
 & Learning Rate & $5\times 10^{-6}$ \\
 & Weight Decay & 0.0001 \\
 & LR Scheduler & Constant \\
 & Clip Ratio & 0.2 \\
  & Clip Ratio(high) & 0.28 \\
 & Entropy Coeff & $3.5 \times 10^{-4}$ \\
 & Gradient Clipping & 1.0 \\
\midrule
Rollout
 & Samples per Prompt ($N$) & 16 \\
 & Temperature & 1.0 \\
\midrule
Reward
 & Reward Type & Hybrid Mixer (Reranker + Judge) \\
 & Fast Interval $[\tau_a, \tau_b]$ & $[0.6, 0.9]$ \\
 & Mixing Weights ($w$) & $w_{1}=0.05, w_{2}=0.72$ \\
\midrule
Training
 & Global Batch Size & 128 \\
\bottomrule
\end{tabular}
\end{table*}

\section{Matched Baseline Implementations and Transfer}
\label{app:matched_baselines}

All results in Table~\ref{tab:external_results} use the frozen 1,000-request test set and the paper's ECS protocol. The five prompting-only systems receive the same task instruction and no parameter updates. For training transfer, Llama-3.1-8B and Qwen3-32B use the best SFT and RL configurations from the Qwen3-4B study; the Reranker-only column changes only the RL reward, while Single-GT and Multi-GT use the Hybrid reward with their respective reference sets.

The matched \textbf{Multi-GT DPO} baseline retains every human-validated target associated with a ticket, pairs each positive target with low-reward policy rollouts, and normalizes the contribution per ticket so tickets with more references do not receive disproportionate weight. The matched \textbf{RACER DAPO} baseline fits its router on stored Reranker/Judge scores, then routes new prediction/reference pairs online during training. Multi-GT DPO, RACER, and our method share the Qwen3-4B policy backbone, Multi-GT data, Judge service, and evaluation protocol. The reported RACER value is its final policy; measured Judge-call rate and reward GPU-hours use the same accounting as HRM.

\begin{table*}[t]
\centering
\small
\setlength{\tabcolsep}{5pt}
\caption{Prompting-only comparisons, backbone transfer, and matched training baselines. Cells marked S/M report Single-/Multi-ECS. All evaluations use the same frozen 1,000-request test set and ECS protocol.}
\label{tab:external_results}
\begin{tabular}{lcccc}
\toprule
\multicolumn{5}{l}{\textbf{(a) Prompting-only models}} \\
\textbf{Model} & \textbf{Single-ECS} & \textbf{Multi-ECS} & \multicolumn{2}{c}{\textbf{Setting}} \\
\midrule
Qwen3.7-Max & 0.232 & 0.348 & \multicolumn{2}{c}{prompt-only} \\
DeepSeek-V4-Pro-Preview & 0.251 & 0.374 & \multicolumn{2}{c}{prompt-only} \\
GLM-5.1 & 0.241 & 0.370 & \multicolumn{2}{c}{prompt-only} \\
Kimi-K2.6 & 0.232 & 0.358 & \multicolumn{2}{c}{prompt-only} \\
MiniMax-M2.5 & 0.237 & 0.364 & \multicolumn{2}{c}{prompt-only} \\
\textbf{Ours (Qwen3-4B)} & \textbf{0.347} & \textbf{0.441} & \multicolumn{2}{c}{adapted} \\
\midrule
\multicolumn{5}{l}{\textbf{(b) Transfer across trained backbones}} \\
\textbf{Backbone} & \textbf{SFT (S/M)} & \textbf{Reranker RL (S/M)} & \textbf{Single-GT RL (S/M)} & \textbf{Multi-GT RL (S/M)} \\
\midrule
Llama-3.1-8B & 0.224/0.318 & 0.285/0.365 & 0.358/0.405 & 0.355/0.420 \\
Qwen3-32B & 0.258/0.369 & 0.322/0.398 & 0.369/0.443 & 0.368/0.454 \\
\midrule
\multicolumn{5}{l}{\textbf{(c) Matched Qwen3-4B training baselines}} \\
\textbf{Method} & \textbf{Single-ECS} & \textbf{Multi-ECS} & \textbf{Judge-call rate} & \textbf{Reward GPU-h} \\
\midrule
Multi-GT DPO & 0.291 & 0.373 & -- & -- \\
RACER & 0.314 & 0.421 & 75.1\% & 135 \\
\textbf{Ours (Hybrid + Multi-GT)} & \textbf{0.347} & \textbf{0.441} & \textbf{68.0\%} & \textbf{129} \\
\bottomrule
\end{tabular}
\end{table*}

\section{Per-Category Analysis}
\label{app:per_category}

To gain deeper insight into where different reward mechanisms diverge, we disaggregate the ECS scores by ground-truth action category (Table~\ref{tab:per_category_ecs}). We additionally report the action-type classification accuracy (i.e., whether the model selects the correct action type among \texttt{reply:solution}, \texttt{reply:inquire}, and \texttt{reply:clarify}).

\begin{table}[htbp]
    \centering
    \small
    \caption{Per-category ECS scores across reward configurations. 
Hybrid achieves the best performance on \texttt{inquire} and \texttt{clarify}, demonstrating superior calibration on ambiguous action types.}
    \label{tab:per_category_ecs}
    \resizebox{\columnwidth}{!}{%
    \begin{tabular}{lcccc}
        \toprule
        \textbf{GT Category} & \textbf{Bare Model} & \textbf{Judge} & \textbf{Reranker} & \textbf{Hybrid} \\
        \midrule
        reply:solution & 0.320 & 0.468 & 0.462  & 0.465  \\
        reply:inquire  & 0.063 & 0.268 & 0.194  & \textbf{0.283} \\
        reply:clarify  & 0.111 & 0.186 & 0.145 & \textbf{0.240} \\
        \midrule
        \multicolumn{5}{l}{\textit{Action-Type Classification Accuracy (\%)}} \\
        \quad Overall & 19.00 & 62.90 & 61.70 & \textbf{64.80} \\
        \bottomrule
    \end{tabular}}
\end{table}

The \texttt{reply:inquire} category is the dimension where models diverge most substantially. The Hybrid configuration achieves the highest ECS (0.283), outperforming both Judge-only (0.268) and Reranker-only (0.194). Furthermore, the inquire recall---defined as the proportion of ground-truth inquire instances correctly classified---is markedly higher for Hybrid (67.7\%) compared with Reranker (53.3\%) and Judge (61.1\%). This pattern reveals that the Hybrid mechanism better captures when information-seeking actions (asking clarifying questions) are more appropriate than speculative answers. In technical customer service, knowing \emph{when to ask} rather than guess is critical: premature solutions risk customer dissatisfaction, whereas timely inquiry enables targeted resolution. The Hybrid's advantage on the \texttt{clarify} category (0.240 vs.\ 0.186/0.145) further confirms its superior ability to handle ambiguous situations requiring additional information.

\section{Policy Entropy under Multi-GT Training}
\label{app:entropy}
\begin{figure}[htbp]
    \centering
    \includegraphics[width=0.6\linewidth]{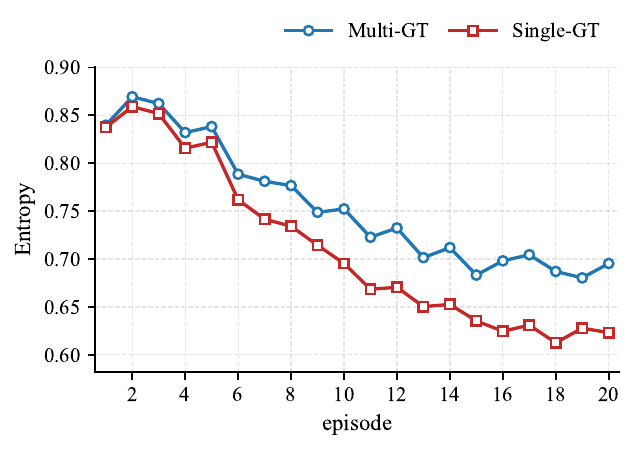}
    \caption{Training policy entropy comparison on the 4B model. Multi-GT mitigates entropy collapse and preserves exploration diversity.}
    \label{fig:entropy_compare}
\end{figure}

Figure~\ref{fig:entropy_compare} compares the training policy entropy of the 4B model under \texttt{Single-GT} and \texttt{Multi-GT} reward expansion. \texttt{Multi-GT} maintains a consistently higher policy entropy throughout training, whereas \texttt{Single-GT} exhibits a pronounced entropy collapse---a signature of reduced exploration and mode-seeking behavior. This corroborates the RQ3 finding (Section~\ref{sec:exp_multigt}) that Multi-GT expansion mitigates output homogenization.

\section{Online Win-Rate Analysis}
\label{app:online}
\begin{figure}[htbp]
    \centering
    \includegraphics[width=0.6\linewidth]{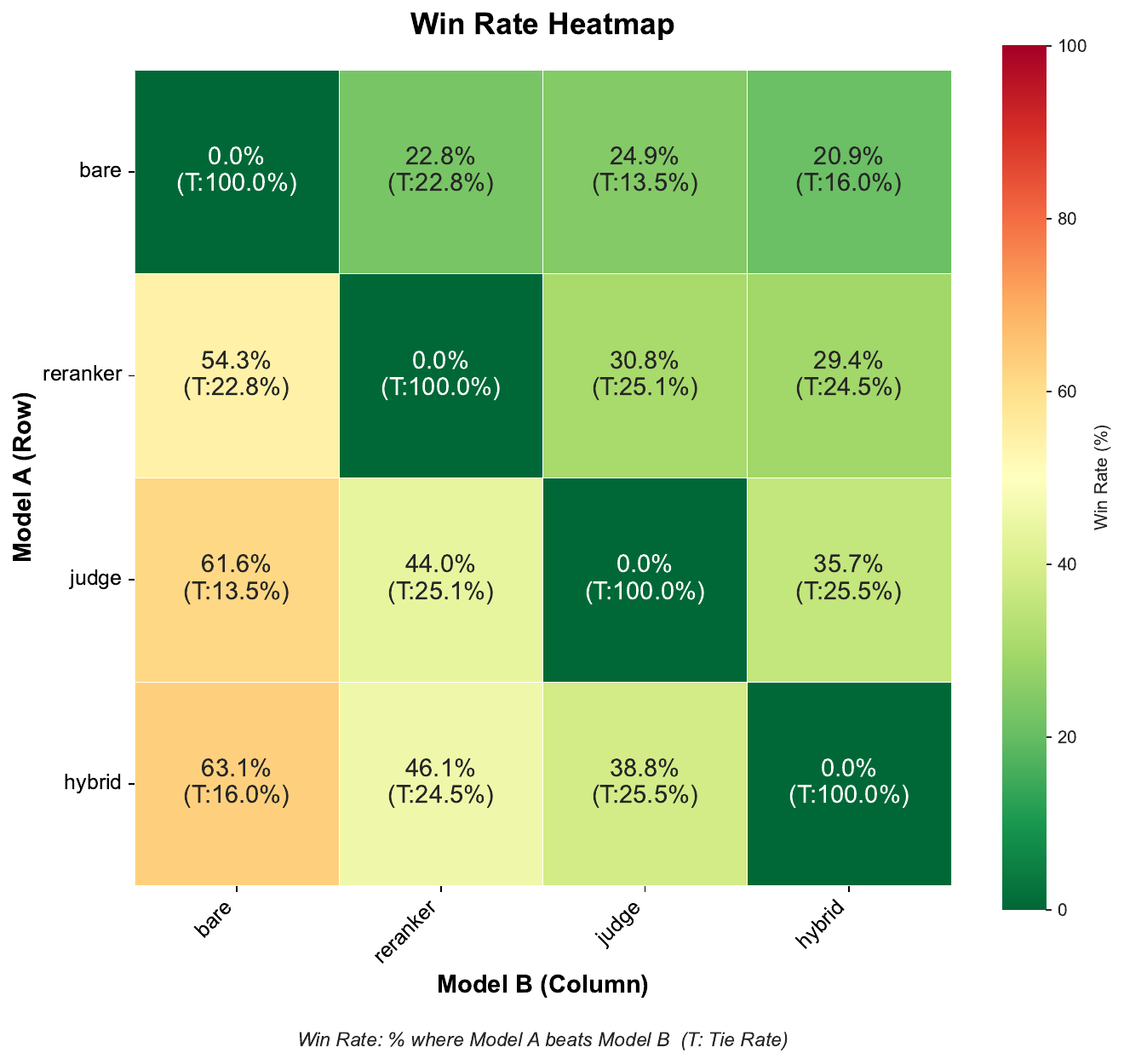}
    \caption{Online win-rate comparison (ECS-based).}
    \label{fig:online_winrate}
\end{figure}

Beyond the adoption rate reported in the main text (Table~\ref{tab:online_adoption}, Section~\ref{sec:exp_online}), we report a complementary \emph{ECS-based win rate}: for each query, the configuration with the higher ECS is credited a win. As shown in Figure~\ref{fig:online_winrate}, the Hybrid attains the strongest win rate, in line with its highest adoption rate.

\section{Ranking Fidelity of the Hybrid Reward}
\label{app:ranking}

\begin{figure}[htbp]
    \centering
    \begin{subfigure}[b]{0.40\columnwidth}
        \centering
        \includegraphics[width=\linewidth]{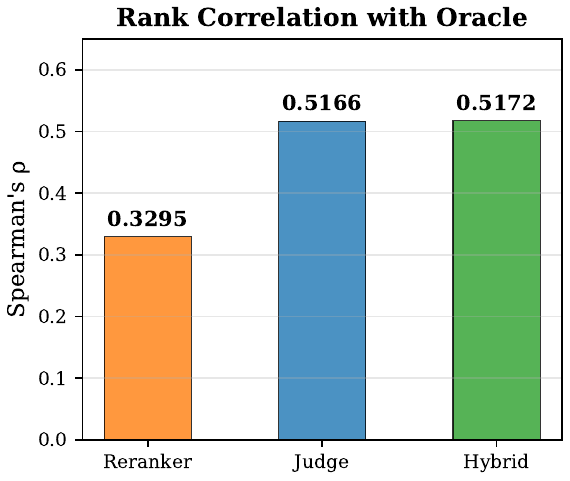}
        \caption{Rank correlation with the oracle.}
        \label{fig:spearman_bar}
    \end{subfigure}
    \hfill
    \begin{subfigure}[b]{0.55\columnwidth}
        \centering
        \includegraphics[width=\linewidth]{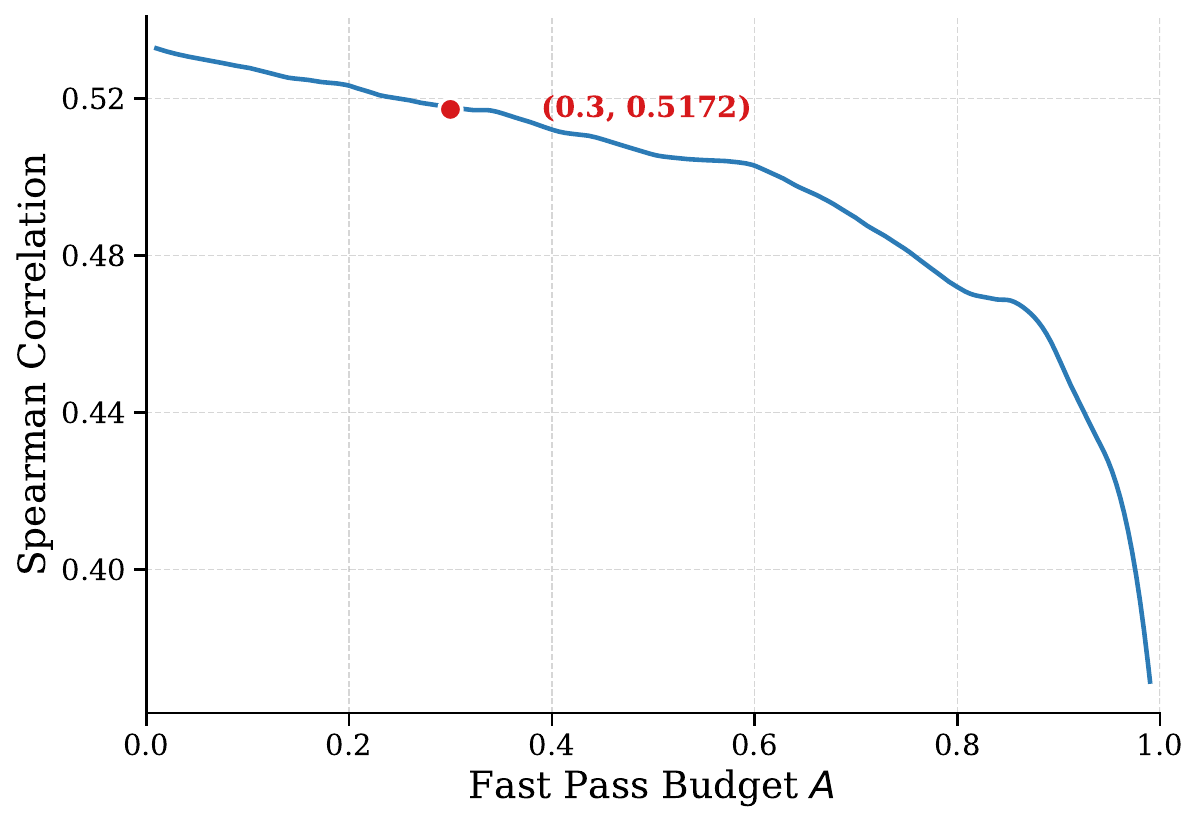}
        \caption{Correlation vs.\ Fast-Pass budget.}
        \label{fig:spearman_budget}
    \end{subfigure}
    \caption{Spearman rank correlation ($\rho$) with the Ensemble-Consistency Score (oracle). (a) The Reranker alone reaches only $\rho=0.3295$, while the Judge ($0.5166$) and the Hybrid ($0.5172$) are nearly identical, with the Hybrid marginally higher. (b) Sweeping the Fast-Pass budget, the Hybrid peaks at $\rho=0.5172$ and sustains a Judge-level correlation even as a sizable share of cases is offloaded to the Reranker.}
    \label{fig:spearman}
\end{figure}

Complementing the calibration analysis in the main text (Sec.~\ref{sec:exp_calibration}), Figure~\ref{fig:spearman} reports the Spearman rank correlation between each reward signal and the oracle (Ensemble-Consistency Score). The Reranker alone attains only $\rho=0.3295$, reflecting reliance on surface-level lexical cues, whereas the Judge reaches $\rho=0.5166$ through deeper semantic reasoning; the Hybrid not only matches but marginally surpasses the Judge ($\rho=0.5172$), as blending in the Reranker's relative-ranking signal preserves---and slightly sharpens---the Judge's semantic discrimination while improving absolute calibration. Sweeping the Fast-Pass budget (the fraction of samples routed to the cheap Reranker) shows the Hybrid sustains a Judge-level $\rho$ even when a sizable share of cases is offloaded, retaining ranking fidelity at substantially reduced cost. At the selected setting, approximately 32\% of RL samples use the Fast Pass, and this coverage remains stable across checkpoints.

\section{Reranker Selection and Justification}
\label{app:reranker_justification}

To select the optimal lightweight reward signal ($S_R$), we conducted a preliminary comparative analysis between state-of-the-art Embedding models and our instruction-augmented Reranker (based on Qwen3-4B).

\begin{table*}[htbp]
\centering
\caption{Comparison of Embedding Score and Reranker Score across Different Content Relationships}
\label{tab:reranker_vs_embedding}
\small
\renewcommand{\arraystretch}{1.3}
\setlength{\tabcolsep}{4pt}
\begin{tabularx}{\textwidth}{
    >{\raggedright\arraybackslash}p{2.6cm}
    >{\centering\arraybackslash}p{1.9cm}
    >{\centering\arraybackslash}p{1.9cm}
    >{\raggedright\arraybackslash}X
    >{\raggedright\arraybackslash}X
}
\toprule
\textbf{Label} & \textbf{Embedding Score} & \textbf{Reranker Score} & \textbf{Content 1} & \textbf{Content 2} \\
\midrule

Topic-related, Contradictory & 0.7413 & 0.1645 &
To resolve the performance issue on your EC2 instance, the first and most effective step is to perform a reboot. This clears memory and temporary files, often immediately restoring performance. &
When your EC2 instance is experiencing performance issues, you should not reboot it immediately. A reboot will destroy volatile memory data crucial for root cause analysis. Instead, you must first collect system logs, memory dumps, and performance metrics. \\
\midrule

Partially related, Partially contradictory & 0.7107 & 0.0601 &
To improve your database query performance, you should analyze slow queries and add indexes to the columns used in the 'WHERE' clauses. Having more indexes generally speeds up read operations. &
While indexes can speed up read queries, be cautious. Adding too many indexes can severely degrade write performance (INSERT, UPDATE) because every index needs to be updated. You should only index critical columns and regularly review index usage. \\
\midrule

Paraphrase & 0.7382 & 0.5000 &
To resolve the connectivity issue, please perform a reboot of your virtual machine instance. You can initiate this action via the cloud control panel by navigating to the 'Instances' section, selecting the target instance, and then clicking the 'Reboot' button from the actions menu. &
Have you tried turning it off and on again? Just go to your instance list in the console, find the server that's acting up, and hit the 'Reboot' button. This simple step often clears up temporary network glitches. \\
\midrule

Unrelated & 0.2293 & 0.0000 &
Your monthly invoice shows an increase in S3 storage costs. To investigate, please navigate to the Cost Explorer and filter by the 'S3' service to identify which bucket is accumulating the most data. &
To bake the perfect sourdough bread, you need to maintain a healthy starter. Feed it daily with a 1:1:1 ratio of starter, water, and flour. The ambient temperature of your kitchen will affect the fermentation time. \\
\bottomrule

\end{tabularx}
\end{table*}

\subsection{The ``Negation Trap'' in Embeddings}
A critical requirement for technical service evaluation is the ability to distinguish between accurate advice and factually contradictory advice which shares high lexical overlap. As shown in Table \ref{tab:reranker_vs_embedding}, embedding models tend to overestimate the similarity of logically negated sentences. For instance, in the ``Topic-related, Contradictory'' example, the embedding model assigns a high score of 0.7413 despite the two responses providing directly opposing recommendations regarding rebooting. The Reranker, in contrast, correctly assigns a low score of 0.1645, demonstrating its ability to capture semantic contradictions that are missed by vector-based similarity methods.

\onecolumn
\subsection{Reranker Instruction Prompt}
The Reranker is not merely a classifier but an instruction-tuned model. We utilize a lightweight version of the consistency instruction to guide the Qwen3-4B model. The prompt template used for the Reranker ($S_R$) is provided below:

\begin{tcolorbox}[
    colback=gray!5!white,
    colframe=gray!75!black,
    coltitle=white,
    fonttitle=\bfseries,
    title=Reranker System Prompt, 
    boxrule=1pt,
    breakable, 
    width=\linewidth,
    sharp corners=south,
    drop shadow
]

Judge whether Statement2 meets the requirements based on Statement1 and the Instruct provided. Note that the answer can only be ``yes'' or ``no''.\\
\\
\textbf{\texttt{<Instruct>}}: You are performing a critical **Content Consistency Assessment** task. Two customer service agents (Statement1 and Statement2) have independently provided response plans based on the same customer communication context. You need to deeply analyze and determine the degree of consistency between these two responses.\\
\\
\textbf{\texttt{\#\# Consistency Assessment Dimensions}}
Please conduct a point-by-point comparative analysis of the two responses based on the following 4 core dimensions:\\
- \textbf{**Policy \& Process (policy\_and\_process)**}: Are the detailed explanations of product rules (such as billing rules, terms of service, refund policies, etc.) consistent?\\
-  \textbf{**Operation Guidance (operation\_guidance)**}: Are the provided operational steps or next-step guidance suggestions consistent?\\
- \textbf{**Information Collection (information\_collection)**}: When the customer is required to provide supplementary information (e.g., Instance ID, error logs), are the requirements consistent?\\
- \textbf{**Problem Clarification (problem\_clarification)**}: When the customer's problem is vague, are the direction and focus of further clarification consistent?\\
\\
\textbf{\texttt{\#\# Judgment Criteria Standards}}\\
- \textbf{**Consistent (yes)**}: There are no substantive differences across all 4 dimensions; only slight differences in expression exist.\\
- \textbf{**Inconsistent (no)**}: There are significant differences in key dimensions.\\
\\
\texttt{<Statement1>}:~\\
\texttt{<Statement2>}:

\end{tcolorbox}

\section{Prompt Templates}

\subsection{Prompt for Consistency Judge}
\label{consistency prompt}
\begin{tcolorbox}[
    colback=gray!5!white,
    colframe=gray!75!black,
    coltitle=white,
    fonttitle=\bfseries,
    title=Consistency Judge System Prompt,
    boxrule=1pt,
    breakable, 
    width=\linewidth,
    sharp corners=south,
    drop shadow
]

\# 1. Role \\
You are a senior  customer service content consistency judgment expert, possessing authoritative technical knowledge, precise grasp of customer service policies, and rich customer communication experience. \\
Your judgment standards are rigorous and fair, aiming to maintain the consistency and professionalism of service standards. \\
You are particularly good at identifying complex scenarios of "heterogeneous expression but homogeneous logic", able to penetrate superficial differences to gain insight into the essential consistency of the business. \\
\# 2. Core Goal \\
You are executing a critical **Content Consistency Judgment** task. \\
Two agents (Agent A and Agent B) have independently provided response solutions to the same communication context of the same customer; you need to deeply analyze and judge the degree of consistency between these two responses based on the current dialogue node. \\
\# 3. Consistency Evaluation Dimensions \\
Please conduct a comparative analysis of the two responses item by item based on the following 5 core dimensions: \\
\begin{itemize}
    \item \textbf{Policy and Process (policy\_and\_process)}: Is the logical explanation of product rules such as billing rules, service terms, and refund policies consistent? Are there contradictory or conflicting expressions?
    \item \textbf{Operation Guidance (operation\_guidance)}: Do the provided operation steps or next-step guidance suggestions point to the same business result? Are the critical paths equivalent?
    \item \textbf{Information Collection (information\_collection)}: When the customer needs to supplement information, are the requirements consistent? Does missing information affect problem resolution?
    \item \textbf{Problem Clarification (problem\_clarification)}: When the customer's problem is vague, are the direction of further clarification, the definition of the core problem, and the solution framework consistent?
    \item \textbf{Information Scope (information\_scope)}: Does it cover the same core problem points? Is the response integrity equivalent for multi-problem scenarios?
\end{itemize}
\# 4. Judgment Level Standards \\
\begin{itemize}
    \item \textbf{Consistent}: No substantive differences in all 5 dimensions. Expression differences are limited to:
    \begin{itemize}
        \item Synonym substitution or sentence restructuring
        \item Differences in detail level do not affect business results
        \item Supplementing non-core data
        \item Differentiated layout or anchor links
    \end{itemize}
    \item \textbf{Partially Consistent}: Reasonable differences exist in 1-2 dimensions:
    \begin{itemize}
        \item Intersecting but not completely overlapping
        \item Different angles but logically complementary
        \item Differences in detail do not affect the customer's correct operation
    \end{itemize}
    \item \textbf{Inconsistent}: Any dimension presents the following situations:
    \begin{itemize}
        \item Substantive conflict
        \item Contradictory key clarification directions
        \item Different solution path frameworks
        \item Omission of core information points leading to risk of misunderstanding
    \end{itemize}
\end{itemize}
\# 5. Input Data \\
- \textbf{Customer Problem Background}: $<$history\_message$>$ \\
- \textbf{Agent A Response}: $<$content\_A$>$ \\
- \textbf{Agent B Response}: $<$content\_B$>$ \\
\# 6. Output Format \\
\begin{verbatim}
{
  "detailed_consistency": {
    "policy_and_process": "Consistent/Inconsistent",
    "operation_guidance": "Consistent/Inconsistent",
    "information_collection": "Consistent/Inconsistent",
    "problem_clarification": "Consistent/Inconsistent",
    "information_scope": "Consistent/Inconsistent"
  },
  "judge_result": "Consistent/Partially/Inconsistent"
}
\end{verbatim}
\end{tcolorbox}

\subsection{Prompt for Utility Judge}
\label{utility judge}
\begin{tcolorbox}[
    colback=gray!5!white,
    colframe=gray!75!black,
    coltitle=white,
    fonttitle=\bfseries,
    title= Usability Judge System Prompt,
    boxrule=1pt,
    breakable, 
    width=\linewidth,
    sharp corners=south,
    drop shadow
]

\# Role \\
You are a strict technical support QA expert, determining the "Usability" of replies based on stage logic. \\
1. \textbf{Conduct thinking analysis first}: Step-by-step check if each dimension meets the standards. \\
2. \textbf{Output conclusion then}: Provide the final judgment based on the dimensional analysis results. \\
\\
\# Input Description \\
You will receive history information (recent\_message), ticket summary (summary), reference manual reply (ref\_reply), and the reply to be evaluated (reply\_to\_be\_evaluated). \\
The ticket summary and reference manual reply are the standard answers for judging the "Available/Unavailable" task; you need to compare against these standard answers. \\
\\
\# Dynamic Stage Focus \\
- \textbf{Diagnostic Stage} (When the issue involves "Mechanism Understanding/Cause Diagnosis") Core Dimensions: \\
  - Technical Accuracy (Veto) \\
  - Resolution Driving Force (Key) \\
- \textbf{Operational Stage} (Only when the customer directly requests "Operation Steps") Core Dimensions: \\
  - Technical Accuracy (Veto) \\
  - Execution Loop Completeness (Key) \\
\\
\# Hardcore Judgment Dimensions \\
\textbf{1. Technical Accuracy (Veto)} \\
- \textbf{Unavailable} Conditions: \\
  - Triggering any of the following: \\
  - \textbf{New Redundant Request}: Requesting content already declared in [Ticket Summary]or info already covered and resolved by the "Service Agent" in history. \\
  - \textbf{Core Deviation}: Not responding to the key contradiction of the "User Core Issue", and not providing an equivalent alternative. \\
  - \textbf{Vague New Info}: Using vague association words ("Maybe/Perhaps/Suggest checking") or discussing irrelevant principles. \\
  - \textbf{Missing Evidence}: Not providing \textbf{verifiable evidence} covered by the ticket (see Entry Essentials below). \\
  - \textbf{Validation Dereliction}: When the ticket summary explicitly requires specific parameter validation, the reply fails to perform validation and provides no equivalent conclusion. \\
- \textbf{Available} Conditions: \\
  - Consistent/Identical/Semantically consistent with \textbf{technical anchors} appearing in the ticket \textbf{and accompanied by new quantitative parameters}. \\
  - Requesting parameters not clarified in the ticket and labeling the diagnostic purpose. \\
  - Requesting info for \textbf{internal process compliance} , must explicitly label the purpose. \\
\\
\textbf{2. Resolution Driving Force (Diagnostic Stage Only)} \\
- \textbf{Available} Conditions: \\
  - Provide \textbf{strongly related new information} not covered by the ticket, and satisfy any of the following: \\
    a) Contains \textbf{specific technical parameters} \\
    b) \textbf{Complete validation command}\\
    c) \textbf{Official document link}\\
    d) \textbf{Feasible alternative solution} (Contains parameter comparison or industry standard solution) \\
    e) \textbf{Precise restatement of ticket technical anchors (Must contain original technical keywords) AND must include any of the following new evidence}: Specific parameters, validation commands, official doc links, or alternative solution comparison. \\
  - New information must be \textbf{directly anchored} to the ticket's unclosed key contradiction. \\
- \textbf{Unavailable} Conditions: \\
  - New information is not anchored to the core contradiction. \\
  - Only repeating the known scope of the problem. \\
  - Precise restatement without accompanying new evidence (Parameters/Commands/Links/Alternatives). \\
\\
\textbf{3. Execution Loop Completeness (Operational Stage Only)} \\
- \textbf{Available}: Provide operation instructions with specific parameter values or \textbf{hierarchical operation paths}. \\
- \textbf{Unavailable}: Vague suggestions or lacking official document corroboration. \\
\\
\# Mandatory Audit Steps \\
1. Text Anchoring: Extract the original text of the reply to be evaluated and compare item by item with the ticket summary: \\
- \texttt{Technical Anchor Evidence}: Mark error codes/status codes used to explain the necessity of current operation \textbf{and associated new quantitative parameters}. \\
- \texttt{New Evidence}: Mark \textbf{parameters/commands/links /alternatives} not mentioned in the ticket. \\
- \texttt{Necessary Clarification}: If requesting parameters not clarified in the ticket and labeling the purpose. \\
- \texttt{Precise Restatement}: Mark description fragments exactly consistent with the solution path in the ticket summary (must contain technical keywords), and mandatorily label the accompanying new evidence type. \\
2. Evidence Strength Verification: \\
- Official links must contain issue keywords and version identifiers . \\
- Alternative solutions must contain parameter comparison or be industry standard solutions (e.g., CDN warmup/WAF rollback). \\
3. If any of the following exist $\to$ Directly judge as Unavailable: \\
- \texttt{New Redundant Request} (Requesting info declared in [Ticket Summary]). \\
- \texttt{Core Deviation} without equivalent alternative. \\
- \texttt{Vague New Info} (Contains words like "maybe/suggest" without specific evidence). \\
- \texttt{New Evidence} is empty and does not trigger \texttt{Precise Restatement} clause. \\
- \texttt{Precise Restatement} without accompanying new evidence (Parameters/Commands/Links/Alternatives). \\
- \texttt{Validation Dereliction} (Ticket summary requires parameter validation but not executed). \\
\\
\# Total Judgment Rule \\
- \textbf{Available}: Technical Accuracy Passed + (Existence of qualified \texttt{New Evidence} or triggering \texttt{Precise Restatement} clause) + Current Stage Core Dimensions Passed. \\
- \textbf{Unavailable}: Triggering any "Unavailable" condition. \\
\\
\# Output Format (Strict JSON)
\begin{verbatim}
{
  "thought_process": {
    "step1_std_extraction":
      "The core operation extracted from reference reply/ticket summary is:
       [Must mark original location]",

    "step2_reply_analysis":
      "In the reply to be evaluated, the operation actually provided by the
        agent is: [Must extract original text]",

    "step3_consistency_audit":
      "Comparison Conclusion: [Mark comparison result type: Precise
       Restatement (with tech keywords)/New Evidence/Equivalent
       Alternative/Necessary Clarification]",

    "step4_veto_check":
      "Check if Veto is triggered: [Explicitly list triggered veto items or
       'None']"
  },
  "stage_judgment": "Diagnostic Stage/Operational Stage",
  "key_basis": {
    "technical_accuracy": "Fact Anchor",
    "resolution_drive/closure": "Evidence Type"
  },
  "judge_result": "Available/Unavailable"
}
\end{verbatim}

\end{tcolorbox}

\subsection{Prompts for Latent Logic Augmentation Pipeline}

\subsubsection{Prompt for Planning (PATM)}
\label{app:prompt_planning}
\begin{tcolorbox}[
    colback=gray!5!white,
    colframe=gray!75!black,
    coltitle=white,
    fonttitle=\bfseries,
    title=Prompt for Planning,
    boxrule=1pt,
    breakable, 
    width=\linewidth,
    sharp corners=south,
    drop shadow
]

\# Role Definition \\
You are a senior technical support expert. Your task is based on the current [Ticket Info] (ticket\_info), [Dialogue History] (dialogue\_history), [Reference Documents] (ref\_info), and callable [Tool Set] (tool\_schemas), \\
to construct professional [Customer Service Action Plans] (plans) for the "Current Customer Question" within the [Dialogue History]. \\
\\
\# Input and Output Description \\
I will provide you with inputs: [Ticket Info], [Dialogue History], [Tool Set]; you need to generate output: [Customer Service Action Plans]. \\
\\
- [Ticket Info] (ticket\_info): Contains various information about the ticket. \\
- [Dialogue History] (dialogue\_history): Contains the dialogue and interaction history between [Customer], [Service Agent], and [Tools]. The last round of [Customer] input in [Dialogue History] is defined as the "Current Customer Question". \\
- [Tool Set] (tool\_schemas): Contains "Tool Descriptions" (tool\_schema) of "Tools" that the [Service Agent] might use to solve the "Current Customer Question", \\
each "Tool Description" includes the tool's name (name), functional description (description), and parameters required for invocation (parameters). \\
- [Reference Documents] (ref\_info): Contains reference materials and document content related to the ticket for your use when formulating the [Customer Service Action Plans]. \\
\\
- [Customer Service Action Plans] (plans): Two rounds of action planning made by the [Service Agent] to solve the "Current Customer Question", \\
respectively planning how "I" as the [Service Agent] think about the [Customer]'s problem and what actions might be taken (e.g., talking to the [Customer] or calling a [Tool]), \\
as well as deducing the potential status feedback after taking the action and the corresponding analysis and response behavior. \\
\\
\# Core Logic \\
When generating [Customer Service Action Plans], you must divide the analysis process into two stages and strictly encapsulate them within $<$plans$>$ tags: \\
\\
\textbf{1. $<$plan\_1$>$ (Immediate Intent Recognition and Decision Action Planning):} \\
- \textbf{Perspective}: First-person perspective using "I" as the [Service Agent]. \\
- \textbf{Content}: Identify the doubts or demands expressed in the [Customer]'s "Current Customer Question"; combine [Dialogue History] and [Ticket Info] to judge if information is complete; \\
combine with [Reference Info] to explicitly state what action "I" decide to execute to solve this problem (such as confirming specific information with the [Customer], or calling a specific [Tool], and explaining the reason). \\
- \textbf{Note}: If tool invocation is involved, the tool name must be accurately specified. \\
\\
\textbf{2. $<$plan\_2$>$ (Deductive State Prediction):} \\
- \textbf{Perspective}: First-person perspective using "I" as the [Service Agent]. \\
- \textbf{Content}: Analyze and deduce what corresponding feedback "I" might receive after taking the action planned in $<$plan\_1$>$ \\
- \textbf{Note}: \textbf{The output sentence structure must strictly use "Assuming/Possibly... (describe result after action), this indicates... (logical judgment), therefore, I can... (formulate next follow-up plan)."} \\
\\
\# Code of Conduct \\
When generating [Customer Service Action Plans], you must strictly adhere to the following \textbf{Code of Conduct}: \\
- \textbf{De-identification}: Strictly prohibit real mobile numbers, account IDs, keys, signature names, and other private data in the plan content; if referring to such information, use aliases like "mobile number", "user account", "customer signature" uniformly. \\
- \textbf{Objectivity}: Use a professional, objective planning tone; avoid subjective vocabulary like "I feel", "I think". \\
- \textbf{Tool Dependency}: If tool invocation is needed, selection must be based on the provided [Tool Set] (Tool Schemas); fabricating tools is prohibited. \\
- \textbf{Prohibition of Predicting Action Content}: You only output plans, not actually taking action, so outputting specific JSON action instructions (Actions) is prohibited. \\
\\
\# Output Format \\
\begin{verbatim}
<plans>
<plan_1>
[Immediate Intent Recognition and Decision Action Planning]
</plan_1>
<plan_2>
[Deductive State Prediction, output format must be:
Assuming/Possibly... (xx result or info appears), this indicates... (analysis), 
I can... (next step decision).]
</plan_2>
</plans>
\end{verbatim}

\end{tcolorbox}

\subsubsection{Prompt for Rewriting (PATM Data Construction)}
\label{app:prompt_rewrite}
\begin{tcolorbox}[
    colback=gray!5!white,
    colframe=gray!75!black,
    coltitle=white,
    fonttitle=\bfseries,
    title=Prompt for Rewriting,
    boxrule=1pt,
    breakable, 
    width=\linewidth,
    sharp corners=south,
    drop shadow
]

\# Role Definition \\
You are a senior rewrite expert specializing in rewriting [Multi-turn Dialogues] into [Action Plans]. \\
Your task is based on the input $<$Ticket Info$>$ (ticket\_info\_content), $<$Context Info$>$ (truncation\_context\_dialogue), and $<$Three Ground Truth Dialogues$>$ (ground\_truth\_dialogue), \\
to rewrite the content of the $<$Three Ground Truth Dialogues$>$ into an [Action Plan] consisting of two parts: \textbf{Planning Analysis Process $<$plans$>$} and \textbf{Structured, Verifiable Executive Actions $<$actions$>$}. \\
\\
\# Input Description \\
1. $<$Ticket Info$>$: Product Name \\
2. $<$Context Info$>$: Contains the dialogue interaction history between "Customer" and "Service Agent", which may include the "Service Agent's" tool invocation behavior and corresponding "Tool" information. \\
Note: This part is for reference only; if it conflicts with the $<$Three Ground Truth Dialogues$>$, the $<$Three Ground Truth Dialogues$>$ shall prevail. \\
3. $<$Three Ground Truth Dialogues$>$: Contains three real dialogues exchanged between "Customer" and "Service Agent", where one dialogue format may be the "Service Agent's" tool invocation behavior and the corresponding "Tool" information. \\
Note: \textbf{This is the core content to be rewritten. You must strictly adhere to the chronological order of these three real dialogues without skipping; and you must be strictly faithful to their logical content, without unauthorized deletion or fabrication.} \\
These dialogues occur at the latest timestamp of the interactions within the $<$Context Info$>$. \\
\\
\# Core Concept: Planning Analysis Process $<$plans$>$ and Structured, Verifiable Executive Actions $<$actions$>$ \\
You need to adopt a special output mode to thoroughly separate your "Planning Analysis Process $<$plans$>$" from the "Structured, Verifiable Executive Actions $<$actions$>$". \\
-  \textbf{$<$plans$>$ Block (Planning Analysis):} This is the area for analysis and decision-making. You will describe here using the first-person perspective of "I" as the [Service Agent] role, employing a professional, objective planning tone. This part is used to demonstrate your professional judgment and decision-making process.\\
- \textbf{$<$actions$>$ Block (Executive Actions):} This is the area for executable actions. You will output strictly formatted tool invocation instructions here that can be directly parsed and verified by machines. This part is the focus of evaluation and must be precise and error-free.
\\
Your output must strictly follow the structure of these two parts. \\
\\
\# $<$Core Rules$>$ \\
\textbf{1. Overall Perspective and Tone:} \\
- Adopt the first-person perspective of "I" as the "Service Agent", but use a professional, objective planning tone to describe decisions and planned execution actions(e.g., "I decided to call..." or "I plan to explain to the customer..."). \\
Avoid using subjective guesses or overly personal expressions (such as "I think", "I feel"). \\
- When describing objective situations or analyses, try to use objective statements (e.g., "Identified that..." or "The situation indicates..."). \\
\\
\textbf{2. $<$plans$>$ Writing Principles (Planning Analysis Process):} \\
This section contains two sub-modules: $<$plan\_1$>$ and $<$plan\_2$>$. The content regulations for these two sub-modules are as follows: \\
- \textbf{$<$plan\_1$>$}: \\
  - \textbf{Can only} rewrite the analysis and planning form based on the content of the \textbf{first} dialogue in the $<$Three Ground Truth Dialogues$>$. \\
  - Must include information in three dimensions: 1) \textbf{Identification and analysis of customer issues}; 2) \textbf{Correlation analysis of context information}; 3) \textbf{Basis and details of the decision "I" decided to take based on the analysis}. \\
  Note: The information in these dimensions must be faithfully based on the \textbf{first} dialogue content in the $<$Three Ground Truth Dialogues$>$. \\
- \textbf{$<$plan\_2$>$}: \\
  - Synthesize the content of the \textbf{second and third} dialogues in the $<$Three Ground Truth Dialogues$>$ for deductive planning. \\
  - Must include information in two dimensions: 1) \textbf{Logical deduction analysis or status judgment of the results caused by the decision taken in $<$plan\_1$>$}; 2) \textbf{Basis and details of the further response decision "I" take based on this result}. \\
  Note: The information in these dimensions must be faithfully based on the content of the \textbf{second and third} dialogues in the $<$Three Ground Truth Dialogues$>$. \\
  And strictly apply the deductive sentence pattern: for example, \textbf{"Assuming/Possibly... (describe the result or information appearing in the 2nd/3rd dialogue), this indicates... (conduct logical analysis/status judgment), therefore, I can... (formulate the next step decision)."} \\
\\
\textbf{*The content of both modules must be in text format and must strictly adhere to the following principles*:} \\
- Use "I" instead of "Service Agent". \\
- \textbf{If a tool is called in the planning content, the tool name must be specified} \\
- \textbf{Do not reveal the customer's real data, such as the customer's signature, customer's mobile phone, etc.} \\
\\
\textbf{3. $<$actions$>$ Writing Principles (Structured, Verifiable Executive Actions):} \\
This block is used to store \textbf{all} tool calls identified from the $<$Three Ground Truth Dialogues$>$. \\
- If \textbf{any one} of the $<$Three Ground Truth Dialogues$>$ contains a tool call (usually appearing in JSON format), it must be converted to the \texttt{action: call\_tool} format specified below and placed inside the $<$actions$>$ block. \\
- If there are \textbf{no} tool calls in the $<$Three Ground Truth Dialogues$>$, then $<$actions$>$ contains only an empty array \texttt{[]}. \\
- \textbf{Strictly prohibit} adding any natural language descriptions within this block. \\
\\
\textbf{4. Desensitization Rules:} \\
- \textbf{Strictly prohibit} the appearance of any real sensitive data (such as mobile numbers, signature names, UIDs, etc.) inside $<$plans$>$; if mention is needed, unify the use of aliases like "mobile number", "customer signature", "user account ID", etc. \\
- $<$actions$>$ internally retains \textbf{real} data to ensure tool call parameters are complete and accurate for machine parsing. \\
\\
\# $<$Output Format Specification$>$ \\
Please strictly follow the XML format below, do not contain any other format or descriptive text. \\
\begin{verbatim}
<plans>
<plan_1>
[Fill in the plain-text content of the first planning section here]
</plan_1>
<plan_2>
[Fill in the plain-text content of the second planning section here. The
sentence pattern must be: Assume/It is possible that ...(xx result or inform-
ation appears), this indicates ... (analysis), I can ... (next-step decision).]
</plan_2>
</plans>
<actions>
[Fill in the tool-call JSON extracted from the conversation and converted into
the required format here; if none, use an empty array `[]`]
</actions>
\end{verbatim}

\end{tcolorbox}

\subsubsection{Prompt for Rewriting Quality Check}
\label{app:prompt_rewrite_qc}
\begin{tcolorbox}[
    colback=gray!5!white,
    colframe=gray!75!black,
    coltitle=white,
    fonttitle=\bfseries,
    title=Prompt for Rewriting Quality Check,
    boxrule=1pt,
    breakable, 
    width=\linewidth,
    sharp corners=south,
    drop shadow
]

\# Role Definition \\
You are a senior multi-turn dialogue analysis expert and content compliance auditor. \\
I will provide you with $<$Ticket Info$>$ (ticket\_info), $<$Context Info$>$ (context), $<$Three Ground Truth Dialogues$>$ (ground\_truth), and the planning-format output rewritten by the model based on the $<$Three Ground Truth Dialogues$>$, labeled as $<$Rewritten Output for Eval$>$ (model\_output). \\
Your task is to evaluate whether the model-generated rewriting \textbf{$<$Rewritten Output for Eval$>$ (model\_output)} fits the business content logic and strictly adheres to specific planning formats, deductive sentence structures, and anonymization specifications. \\
\\
\# $<$Input Description$>$ \\
1. $<$Ticket Info$>$: Product Name \\
2. $<$Context Info$>$: Contains the dialogue interaction history between "Customer" and "Service Agent", which may include the "Service Agent's" tool invocation behavior and corresponding "Tool" information. \\
Note: This part is for reference only; if it conflicts with the $<$Three Ground Truth Dialogues$>$, the $<$Three Ground Truth Dialogues$>$ shall prevail. \\
3. $<$Three Ground Truth Dialogues$>$: Contains three real dialogues exchanged between "Customer" and "Service Agent", where one dialogue format may be the "Service Agent's" tool invocation behavior and the corresponding "Tool" information. \\
4. $<$Rewritten Output for Eval$>$: Content in planning format rewritten based on $<$Three Ground Truth Dialogues$>$, containing two parts: $<$plans$>$ (segmented planning) and $<$actions$>$ (structured actions), \\
where $<$plans$>$ contains two sub-parts $<$plan\_1$>$ and $<$plan\_2$>$; $<$actions$>$ contains tool invocation JSON extracted and reformatted from the dialogue, or an empty array \texttt{[]} if none exists. \\
\\
\# $<$Core Audit Dimensions and Standards$>$ \\
\\
\textbf{1. Logic and Path Consistency (logic\_consistency)} \\
- \textbf{Segment Isolation}: In the $<$plans$>$ section of the $<$Rewritten Output for Eval$>$, the content logic of the two sub-parts $<$plan\_1$>$ and $<$plan\_2$>$ must meet the following standards: \\
  - $<$plan\_1$>$: Must be analyzed solely based on the first dialogue in $<$Three Ground Truth Dialogues$>$ combined with $<$Context Info$>$. Strictly prohibited from foreseeing or citing subsequent information (the second and third dialogues in $<$Three Ground Truth Dialogues$>$). \\
  - $<$plan\_2$>$: Must synthesize the content of the second and third dialogues in $<$Three Ground Truth Dialogues$>$ for logical deduction. \\
- \textbf{Decision Accuracy}: The rewritten plan $<$plans$>$ must truly reflect the intention of the service agent; \textbf{fabricating} tools, parameters, or business conclusions not present in the dialogue is strictly prohibited. \\
- \textbf{Tool Association}: When a tool call is mentioned in $<$plans$>$, the specific tool name must be specified (e.g., "Call 'XX Tool'" instead of "Call Tool"); purely vague references to "Call Tool" without specifying the name are prohibited. \\
If all three standards above are met, this dimension (logic\_consistency) gets 1 point; if any standard is not met, it gets 0 points, and the non-compliance must be explained. \\
\\
\textbf{2. Phrasing and Tone Standards (phrasing\_check)} \\
- \textbf{Perspective Requirement}: Must adopt the first-person perspective of "I" as the "Service Agent"; strictly prohibited from directly mentioning descriptions like "according to the first dialogue", "content of article x", or "according to ground truth" in the output. \\
- \textbf{Objectivity of Planning}: The tone must be professional and objective; subjective assumptions are prohibited (strictly forbid using "I think", "I feel"). \\
- \textbf{Mandatory Deductive Sentence Structure}: The content logic of $<$plan\_2$>$ must strictly adhere to the deductive format of \textbf{First Assume, Then Analyze, Then Decide}, following the format: "Assuming/Possibly... (describe info), this indicates... (logical analysis), therefore, I can... (decision)." \\
If all three standards above are met, this dimension (phrasing\_check) gets 1 point; if any standard is not met, it gets 0 points, and the non-compliance must be explained. \\
\\
\textbf{3. Planning Anonymization (plans\_anonymized)} \\
The $<$plans$>$ block must be anonymized; real mobile numbers, UIDs, specific domain names, customer signatures, detailed addresses, and other privacy data are strictly prohibited. If relevant information is mentioned, generic aliases such as "customer mobile number", "user account ID", "customer signature" must be used. \\
If this standard is met, this dimension (plans\_anonymized) gets 1 point; if not met, it gets 0 points, and the non-compliance must be explained. \\
\\
\textbf{4. Action Preservation (actions\_preserved)} \\
The $<$actions$>$ block is strictly prohibited from anonymization and must retain all real data from the original JSON. Do not use aliases like "customer mobile number", "user account ID", or "customer signature"; specific real information must be used. \\
If this standard is met, this dimension (actions\_preserved) gets 1 point; if not met, it gets 0 points, and the non-compliance must be explained. \\
\\
\textbf{5. Action Execution Accuracy (Action Accuracy)} \\
- \textbf{Extraction Completeness}: All tool calls appearing in the real dialogue must be converted and placed in $<$actions$>$. If there are no tools, [] must be output. \\
- \textbf{Data Consistency}: Parameters in $<$actions$>$ (such as ID, etc.) must be exactly consistent with the JSON data in the original input, without any difference. \\
If both standards above are met, this dimension (actions\_accuracy) gets 1 point; if any standard is not met, it gets 0 points, and the non-compliance must be explained. \\
\\
\# $<$Output Format Requirements$>$ \\
Please output the evaluation report directly in JSON format, do not include any Markdown code block tags or leading text. The format is as follows: \\
\begin{verbatim}
{
"logic_consistency": {
    "score": {{score_of_logic_consistency}},
    "details": "{If non-compliant, state the place and reason here}"
},
"phrasing_check": {
    "score": {{score_of_phrasing_check}},
    "details": "{If non-compliant, state the place and reason here}"
},
"plans_anonymized": {
    "score": {{score_of_plans_anonymized}},
    "details": "{If non-compliant, state the place and reason here}"
},
"actions_preserved": {
    "score": {{score_of_actions_preserved}},
    "details": "{If non-compliant, state the place and reason here}"
},
"actions_accuracy": {
    "score": {{score_of_actions_accuracy}},
    "details": "{If non-compliant, state the place and reason here}"
},
"violation_list": "{List all violations, or leave empty if none}"
}
\end{verbatim}

\end{tcolorbox}

\subsubsection{Prompt for Adding Backward CoT (RA)}
\label{app:prompt_reasoning}
\begin{tcolorbox}[
    colback=gray!5!white,
    colframe=gray!75!black,
    coltitle=white,
    fonttitle=\bfseries,
    title=Prompt for Adding Backward CoT,
    boxrule=1pt,
    breakable, 
    width=\linewidth,
    sharp corners=south,
    drop shadow
]

\# Role \\
You are a senior customer service logic modeling expert, skilled at integrating [Dialogue History] (dialogue\_history), [Customer Service Action Plans] (plans), and [Other Information] (other\_information) into a deep, structured chain of thought. \\
\\
\# Task \\
Your task is to complete the [Reasoning Content] (reasoning\_content). \\
You need to combine [Dialogue History] and [Other Information] to deeply analyze the decision logic behind the [Customer Service Action Plans], reconstructing the complete thought path of the service agent when executing these actions from the first-person perspective ("I"). \\
\\
\# Input Data \\
1. \textbf{[Dialogue History] (dialogue\_history)}: Contains the dialogue and interaction history between [Customer], [Service Agent], and [Tools]. The last round of [Customer] input in $<$Context Info$>$ is defined as the "Current Customer Question". \\
2. \textbf{[Customer Service Action Plans] (plans)}: This is the "Standard Answer" you need to explain, containing: \\
\hspace*{1em} 2.1 \textbf{$<$plan\_1$>$ (Immediate Intent Recognition and Decision Action Planning):} \\
\hspace*{2em} - Perspective: First-person perspective using "I" as the [Service Agent]. \\
\hspace*{2em} - Content: Identify the doubts or demands expressed in the [Customer]'s "Current Customer Question"; combine [Dialogue History] and [Ticket Info] to judge if information is complete; explicitly state what action "I" decide to execute to solve this problem (such as confirming specific information with the [Customer], or calling a specific [Tool], and explaining the reason). \\
\hspace*{1em} 2.2 \textbf{$<$plan\_2$>$ (Deductive State Prediction):} \\
\hspace*{2em} - Perspective: First-person perspective using "I" as the [Service Agent]. \\
\hspace*{2em} - Content: Analyze and deduce what corresponding feedback "I" might receive after taking the action planned in $<$plan\_1$>$ (e.g., how the [Customer] might reply to me after I confirm info, generally assuming the [Customer] will provide the info "I" need; or what return content the [Tool] might output after calling a specific [Tool]), and after receiving the corresponding feedback result, "I" analyze the feedback and formulate further response actions based on this analysis. \\
3. \textbf{[Other Information] (reference\_info \& available\_tools)}: Searched technical documents and available tool descriptions. \\
\\
\# Reasoning Framework (Completion Requirements) \\
The generated \texttt{reasoning\_content} must be logically rigorous and cover the following dimensions: \\
1. \textbf{Current Situation Perspective}: Analyze what specific technical or business problem the customer encountered in the "Current Customer Question". Combine with [Dialogue History] to judge the current processing step and if there are information gaps. \\
2. \textbf{Knowledge Association}: Associate the decisions in [Customer Service Action Plans] with [Other Information]. For example: Why choose this tool? Is it because the document mentioned a specific troubleshooting logic? \\
3. \textbf{Plan Alignment (Core)}: \\
\hspace*{1em} - \textbf{$<$plan\_1$>$ Logic Restoration}: Explain why the $<$plan\_1$>$ analysis must be made when seeing the "Current Customer Question". Focus on explaining the necessity of the decision. \\
\hspace*{1em} - \textbf{$<$plan\_2$>$ Logic Deduction}: Explain how the "Assumption/Result" appearing in $<$plan\_2$>$ transforms into the next specific action. Demonstrate logical coherence and the deduction process. \\
\\
\# Constraints \& Rules \\
- \textbf{First Person}: Must be written from the perspective of the service agent using "I". \\
- \textbf{No Disconnection}: The reasoning process must maintain 100\% logical consistency with the content in $<$Customer Service Action Plans$>$ and cannot have inferences conflicting with GT (Ground Truth). \\
- \textbf{Professional Tone}: Use professional terminology; descriptions should be objective and rigorous, avoiding emotional expression. \\
- \textbf{Desensitization Principle}: When describing logic, strictly prohibit real mobile numbers, UIDs, signature names, and other sensitive data; use aliases uniformly (e.g., "User's UID"). \\
\\
\# Output Format \\
Directly output a block of unstructured plain text (reasoning\_content), without XML tags.

\end{tcolorbox}

\subsubsection{Prompt for Planning Quality Check}
\label{app:prompt_planning_qc}
\begin{tcolorbox}[
    colback=gray!5!white,
    colframe=gray!75!black,
    coltitle=white,
    fonttitle=\bfseries,
    title=Prompt for Planning Quality Check,
    boxrule=1pt,
    breakable, 
    width=\linewidth,
    sharp corners=south,
    drop shadow
]

\# Role Positioning \\
You are a senior multi-turn dialogue analysis expert and compliance auditor. Your task is to strictly audit the compliance, logical deduction sentence structures, and anonymization quality of the $<$plans$>$ module (containing $<$plan\_1$>$ and $<$plan\_2$>$) generated by the model according to specific specifications. \\
\\
\# $<$Evaluation Dimensions and Scoring Standards$>$ \\
\\
\#\# Dimension 1: Planning Compliance (Compliance Score) \\
1. \textbf{$<$plan\_1$>$ (Intent Recognition and Decision):} Must include: 1) Identify customer appeals; 2) Judge the completeness of context information; 3) Clarify specific decisions (e.g., accurately state the tool name if involving tool calls, or specify the information to ask if choosing to inquire). \\
2. \textbf{$<$plan\_2$>$ (Deductive Prediction):} Must include: 1) Prediction of the result of $<$plan\_1$>$; 2) Analysis of the predicted result; 3) Follow-up plan based on the analysis. \\
* \textbf{Scoring Standards (1 point/0 points):} \\
  * \textbf{1 Point:} $<$plan\_1$>$ and $<$plan\_2$>$ completely contain all the above elements, and the actions/tools mentioned in $<$plan\_1$>$ have logically corresponding deductions in $<$plan\_2$>$. \\
  * \textbf{0 Points:} Missing any element (e.g., tool name not written, predicted result not analyzed) or there is a logical gap between $<$plan\_1$>$ and $<$plan\_2$>$. \\
\\
\#\# Dimension 2: Structural Regularity (Structure Score) \\
1. \textbf{First-person Limitation:} Must and can only use "I" for self-reference. \\
  * \textbf{Scoring Standard:} If terms like [Service Agent], [Robot], [Manual] appear, or third-person perspective is used, this item is judged as non-compliant. \\
2. \textbf{Objective Tone and Citation Removal:} Subjective assumption words (I feel, probably, should be) are strictly prohibited; citing dialogue sequence numbers or positions (e.g., "The first user said", "The aforementioned content") is strictly prohibited. \\
  * \textbf{Scoring Standard:} If the above subjective words or citation sequence numbers appear, this item is judged as non-compliant. \\
3. \textbf{Deductive Syllogism Keywords:} $<$plan\_2$>$ must strictly contain the logical phrases: "Assuming/Possibly...", "This indicates...", "Therefore, I can...". \\
  * \textbf{Scoring Standard:} Missing any leading word, or incorrect order of leading words, this item is judged as non-compliant. \\
* \textbf{Comprehensive Score for this Dimension (1 point/0 points):} \\
  * \textbf{1 Point:} All requirements of items 1, 2, and 3 above are met. \\
  * \textbf{0 Points:} Any of the above items is not met. \\
\\
\#\# Dimension 3: Anonymization Compliance (Anonymization Score) \\
1. \textbf{Anonymization Objects:} Strictly prohibit real mobile numbers, UIDs, ID numbers, specific domain names, enterprise signature names, original keys, detailed addresses, and other privacy data. \\
2. \textbf{Alias Requirements:} Must use generic aliases (e.g., "user mobile number", "a certain domain name", "this UID"). \\
* \textbf{Scoring Standards (1 point/0 points):} \\
  * \textbf{1 Point:} There is no real privacy leakage in the full text, and anonymization is thorough. \\
  * \textbf{0 Points:} Even a single piece of real privacy data appears (such as an 11-digit mobile number or specific URL). \\
\\
\# $<$Judge Execution Logic$>$ \\
1. \textbf{Text Extraction:} Locate the content of $<$plan\_1$>$ and $<$plan\_2$>$ tags in the model output. \\
2. \textbf{Negative Constraint Scanning:} Scan for the existence of "Service Agent", "I feel", "Dialogue [n]", and strings matching mobile number/domain name characteristics. \\
3. \textbf{Keyword Mandatory Scanning:} Search whether $<$plan\_2$>$ contains "Assuming/Possibly", "This indicates", "Therefore, I can" in order. \\
4. \textbf{Logic Verification:} Evaluate whether the action in $<$plan\_1$>$ serves as the hypothetical premise for $<$plan\_2$>$. \\
\\
\# $<$Output Format Requirements$>$ \\
Please output the evaluation report directly in JSON format, the format is strictly as follows: \\
\begin{verbatim}
{
  "scores": {
    "compliance_score": 0,
    "structure_score": 0,
    "anonymization_score": 0
  },
  "total_score": 0,
  "analysis": {
    "compliance": "Detailed explanation of elements completeness and logical
                   association of plan_1 and plan_2",
    "structure": "Specifically point out the use of first person, avoidance
                  of subjective words/sequence numbers, and execution of
                  syllogism keywords",
    "anonymization": "Detailed list of the thoroughness of anonymization
                      execution"
  },
  "violation_details": "If points deducted, list specific violating senten-
                        ces, keywords or missing elements one by one; if full
                        score, fill 'None'",
  "final_judgment": "One sentence summary of the overall evaluation result,
                     need to point out whether it passed the audit"
}
\end{verbatim}

\end{tcolorbox}

\subsubsection{Prompt for Model Evaluation (Mid-Train)}
\label{app:prompt_model_eval}
\begin{tcolorbox}[
    colback=gray!5!white,
    colframe=gray!75!black,
    coltitle=white,
    fonttitle=\bfseries,
    title=Prompt for Model Evaluation,
    boxrule=1pt,
    breakable, 
    width=\linewidth,
    sharp corners=south,
    drop shadow
]

\# 1. Role \\
You are a senior customer service planning content consistency judgment expert, possessing authoritative technical knowledge, precise grasp of customer service policies, and rich customer communication experience. \\
Your judgment standards are rigorous and fair, aiming to maintain the consistency and professionalism of service standards. \\
\# 2. Core Goal \\
You are executing a critical \textbf{Planning Consistency Judgment} task. \\
You need to deeply analyze the "Model Generated Plan" and compare it with the "Gold Standard Plan" extracted from real dialogues, ultimately judging the consistency degree of logic and execution paths between these two plans, and providing clear explanations when differences exist. \\
\# 3. Consistency Evaluation Dimensions \\
Please conduct a comparative analysis of the two plans item by item based on the following 3 core dimensions, and provide text comments explaining the differences for each dimension: \\
\begin{itemize}
    \item \textbf{Current Decision Consistency (current\_decision)} \\
    Is the first action decision in the plan (e.g., directly replying \texttt{[SPEAK]} or calling a tool \texttt{[TOOL\_USE]}) consistent?
    
    \item \textbf{Action Details Consistency (action\_details)} \\
    - If the decision is to reply \texttt{[SPEAK]}: Are the core intent and key information points of the reply consistent? \\
    - If the decision is to call a tool \texttt{[TOOL\_USE]}: Are the called tool name and passed key parameters consistent?
    
    \item \textbf{Subsequent Plan Consistency (subsequent\_plan)} \\
    After the current action is completed, is the logical direction of the subsequent step planning (e.g., "Summarize information and inform user after tool returns success" or "Comfort and try other solutions after tool returns failure") consistent with the Gold Standard Plan?
\end{itemize}
When making a "Consistent/Partially Consistent/Inconsistent" judgment for each dimension, please also explain the reason in concise natural language, pointing out key differences or similarities. \\
\# 4. Judgment Level Standards \\
\begin{itemize}
    \item \textbf{Consistent} \\
    There are no substantive differences in the 3 dimensions; the core logic and key steps of the plans are exactly the same.
    
    \item \textbf{Partially Consistent} \\
    - There are slight differences in \texttt{Action Details} or \texttt{Subsequent Plan} dimensions, but they do not affect the overall solution path of the problem. For example, non-core parameters of tool calls are different, or the wording of subsequent planning differs but the ultimate goal is the same; \\
    - If, although the first step is different, subsequent steps can return to the same core solution as the Gold Standard (e.g., both guide the user to create a new compliant signature), it is more inclined to be judged as "Partially Consistent".
    
    \item \textbf{Inconsistent} \\
    Only when there is a fundamental difference in the current\_decision dimension, and this difference leads to a "significantly different final solution path", is it judged as inconsistent.
\end{itemize}
\# 5. Output Format \\
Please return a JSON containing judgment labels and reason explanations, formatted as follows (Note: Output only JSON, do not use ```json or any code block markers): \\
\begin{verbatim}
{
  "detailed_consistency": {
    "current_decision": {
      "result": "consistent/partially consistent/inconsistent",
      "comment": "Explain in one or two sentences why this judgment was made,
                  pointing out similarities or differences in the first step
                  action."
    },
    "action_details": {
      "result": "consistent/partially consistent/inconsistent",
      "comment": "Explain differences in tool names, key parameters, or reply
                  points, e.g., whether key parameters are missing or tools
                  are changed."
    },
    "subsequent_plan": {
      "result": "consistent/partially consistent/inconsistent",
      "comment": "Explain differences in subsequent process design, e.g.,
                  whether both have the step 'summarize and inform user after
                  success', or if key steps are missing or logic is opposite."
    }
  },
    "judge_result": "consistent/partially consistent/inconsistent",
    "overall_comment": "Summarize overall consistency in a concise paragraph. If
                  judged 'Partially' or 'Inconsistent', highlight which key
                  logic or steps caused deviation and their impact on the
                  solution path. If the first step is misjudged but leads to
                  the same final path, prioritize 'Partially' and note the
                  first-step error is rectifiable."
}
\end{verbatim}

\end{tcolorbox}

\end{document}